%% file: completePaper.tex
\pdfoutput=1 

\documentclass[10pt]{article}
\usepackage{relsize}

\usepackage{color}
\usepackage{listings}
\usepackage{graphicx}
\usepackage{subfig}
\usepackage{epstopdf}
\usepackage{array}
\usepackage{multirow}
\usepackage{amsmath}

\usepackage[labelsep=period]{caption}

\usepackage[OE]{express}

\definecolor{red}{rgb}{0.8,0,0}
\definecolor{darkred}{rgb}{0.6,0,0}
\definecolor{green}{rgb}{0.0,0.5,0}
\definecolor{blue}{rgb}{0,0,0.75}
\definecolor{orange}{rgb}{1,0.6,0.2}
\definecolor{purple}{rgb}{0.7,0.0,0.7}
\definecolor{cyan}{rgb}{0.0,0.7,0.7}

\newcommand{\twopartdef}[3]
{
	\left\{
		\begin{array}{ll}
			#1 & \mbox{if } #2 \\
			#3 & \mbox{elsewhere} 
		\end{array}
	\right.
}

\newcommand{\new}[1]{#1}

\newcommand{\shortcite}[1]{\cite{#1}}

\newcommand{\Fig}[1]{Fig.~\ref{fig:#1}}
\newcommand{\FigFull}[1]{Figure~\ref{fig:#1}}

\newcommand{\Sec}[1]{Section~\ref{sec:#1}}

\newcommand{\Apx}[1]{Appendix~\ref{ap:#1}}

\newcommand{\diff}[1]{d{#1}}

\newcommand{\pdf}{p}
\newcommand{\point}[1]{\mathbf{#1}}
\newcommand{\domain}[1]{\mathbf{{#1}}}
\newcommand{\sMatrix}[1]{\mathbf{{#1}}}

\newcommand{\discretization}[1]{\bar{#1}}

\newcommand{\sPoint}{x}
\newcommand{\sVisPoint}{p}
\newcommand{\sLight}{s}

\newcommand{\px}{\point{\sPoint}}
\newcommand{\vispoint}{\point{\sVisPoint}}
\newcommand{\pixel}{\discretization{\vispoint}}
\newcommand{\light}{\point{\sLight}}

\newcommand{\Visible}{\domain{P}}
\newcommand{\Lights}{\domain{S}}
\newcommand{\Scene}{\domain{X}}
\newcommand{\Pixels}{\discretization{\Visible}}
\newcommand{\Voxels}{\discretization{\Scene}}

\newcommand{\NumLights}{S}
\newcommand{\NumPixels}{\discretization{P}}
\newcommand{\NumVoxels}{\discretization{X}}
\newcommand{\NumFrames}{T}
\newcommand{\NumTriangles}{\mathcal{T}}

\newcommand{\sT}{t}
\newcommand{\Signal}{I}
\newcommand{\Rad}{L}
\newcommand{\Rado}{\Rad_o}
\newcommand{\sSOL}{c}

\newcommand{\fProjection}{\psi}
\newcommand{\fInvProjection}{\fProjection^{-1}}
\newcommand{\fOptDepth}{\tau}
\newcommand{\fGeom}{G}
\newcommand{\fVis}{V}
\newcommand{\fBRDF}{f}
\newcommand{\sElipsoid}{E}

\newcommand{\sTessLevel}{o}
\newcommand{\sTessSpheres}{\domain{O}}
\newcommand{\AreaTriangle}{\alpha}
\newcommand{\sEllipsoidTMatrix}{\sMatrix{M}}

		\newcommand{\ra}{\!\rightarrow\!}

		\newcommand{\approptoinn}[2]{\mathrel{\vcenter{\offinterlineskip\halign{
				\hfil$##$\cr#1\propto\cr\noalign{\kern2pt}#1\sim\cr\noalign{\kern-2pt}
		}}}}

		\newcommand{\Real}{\mathbb{R}}
		\newcommand{\Order}[1]{O(#1)}		

\begin{document}

\title{Fast back-projection for non-line of sight reconstruction}
\vspace{-0.15cm}

\author{Victor Arellano, Diego Gutierrez, and Adrian Jarabo\authormark{*}}

\vspace{-0.05cm}
\address{Universidad de Zaragoza - I3A, Zaragoza 50018, Spain}

\email{\authormark{*}ajarabo@unizar.es} 

\vspace{-0.2cm}
\begin{abstract}
Recent works have demonstrated non-line of sight (NLOS) reconstruction by using the time-resolved signal from multiply scattered light. These works combine ultrafast imaging systems with computation, which back-projects the recorded space-time signal to build a probabilistic map of the hidden geometry. Unfortunately, this computation is slow, becoming a bottleneck as the imaging technology improves. In this work, we propose a new back-projection technique for NLOS reconstruction, which is up to \textit{a thousand times} faster than previous work, with almost no quality loss. 
We base on the observation that the hidden geometry probability map can be built as the intersection of the three-bounce space-time manifolds defined by the light illuminating the hidden geometry and the visible point receiving the scattered light from such hidden geometry. This allows us to pose the reconstruction of the hidden geometry as the voxelization of these space-time manifolds, which has lower theoretic complexity and is easily implementable in the GPU. We demonstrate the efficiency and quality of our technique compared against previous methods in both captured and synthetic data. 
%
\end{abstract}
\linespread{0.95}

\bibliographystyle{osajnl}

\section{Introduction}
\label{sec:intro}

One of the core applications of time-resolved imaging (see e.g.~\cite{Jarabo2017,Bhandari2016} for recent surveys on the field) is the capability to robustly capture depth from a scene, by being able to track the time of arrival of photons. Geometry reconstruction techniques have significantly benefited from this, but in the last years the applicability of time-resolved imaging has gone beyond directly visible geometry, to include non-line of sight (NLOS) imaging~\cite{Velten2012nc}. This technique allows reconstructing occluded objects by analyzing multiple-scattered light, even in the presence of turbid media~\cite{Han2000,Raviv2014,Heide2014scattering}.

The first prototype of this technology was demonstrated with a femtosecond laser and a streak camera~\cite{Velten2013}.
Further work explored alternative hardware setups, including correlation-based time-of-flight cameras~\cite{Heide2014mirrors}, single photon avalanche diodes (SPAD)~\cite{Buttafava2015}, laser-gated sensors~\cite{Laurenzis2014}, or even common DSLR cameras~\cite{Klein2016}. These setups are cheaper and more portable, although at the cost of sacrificing time resolution. 

Despite all these available hardware options, the \textit{reconstruction} step is still a bottleneck, limiting the applicability of this technology in the wild. Different approaches have been proposed for reconstruction, either by solving a non-convex optimization on three-dimensional geometry~\cite{Hullin2014,Klein2016} or a depth map~\cite{Heide2014mirrors}, or by back-projecting the space-time captured image on a voxelized geometry representation~\cite{Velten2012nc,Gupta2012}. In both cases, the large amount of data being processed, together with the complexity of the reconstruction algorithms, impose computation times in the order of several hours. 

Back-projection reconstruction builds a 3D probability map based on the recorded time-resolved image, encoding the time of arrival of photons reflected by the hidden geometry. It exploits the correlation between the time of arrival of a photon and the total distance traveled, back-projecting this distance to the reconstruction volume. 
This approach has several advantages over the alternative methods: First, it avoids the need to solve a non-convex optimization problem, which may fall in local minima. Second, the memory cost is significantly lower.
 And third, it is relatively robust to noisy captures, which is particularly important when aiming for low capturing times. 
However, current back-projection methods are very inefficient and redundant, which results in a poor scalability with the resolution of the voxelized scene. This imposes a hard limit of the quality of the reconstruction, regardless of the used imaging technology. 

\begin{figure}[t]
  \centering
  	\def\svgwidth{0.7\textwidth}\footnotesize
    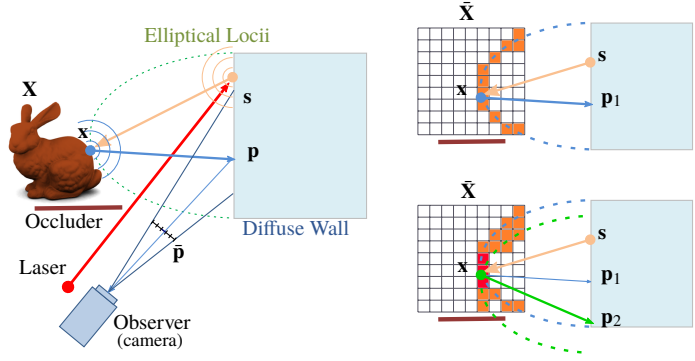
  \caption{Overview of our method. Left: Illustration of our reconstruction setup. A laser pulse is emitted towards a diffuse wall, creating a virtual point light $\light$ illuminating the occluded scene. The reflection of the occluded geometry travels back to the diffuse wall, which is imaged by the camera. The total propagation time from a hidden surface point $\px$ forms an ellipsoid with focal points at $\light$ and $\vispoint$. Right: The intersection of several of these ellipsoids defines a probability map for the occluded geometry, from which the reconstruction is performed with a speed-up factor of three orders of magnitude over previous approaches.
  }
  \label{fig:scene}
\end{figure}

In this work we propose a new back-projection reconstruction method that yields a speed-up factor of three orders of magnitude over previous NLOS reconstruction approaches, thus addressing the main pending issue limiting the applicability of recent approaches. 
We build on the key observation that the manifold of probably points that might reflect light towards an observed point at a certain time form an ellipsoid with poles at the light source and the observed point, and with radii the time of flight of photons (\Fig{scene}). This allows us to build the probability map as the intersection of the ellipsoids of multiple space-time measurements. Posing the problem in this ellipsoid space allows to avoid redundancy, which reduces the theoretical complexity of the algorithm to just the number of space-time measurements. 

More importantly, formulating the problem this way is equivalent to performing voxelization of the ellipsoids geometry, which is very suitable for modern GPUs.

This combination of lower computational complexity and an efficient hardware-accelerated implementation allows us to increase efficiency further: our techniques allows computing the probability maps of large datasets in the order of seconds with a minimum increase in error, while on the other hand allows significantly higher-resolution reconstructions with a negligible added cost. We demonstrate these capabilities by evaluating our method using both synthetic and real captured data, and comparing with existing approaches. In addition, we make our code publicly available at \url{http://giga.cps.unizar.es/~ajarabo/pubs/nlosbackprojectionOExp17/code/}. 

\section{Back-projection for NLOS reconstruction}
\label{sec:scene}
The goal of NLOS reconstruction methods is to recover an unknown hidden scene $\Scene\in\Real^3$ from the measurements on visible known geometry $\Visible\in\Real^3$. Such $\Visible$ is defined by a bijective map $\fProjection: \Real^2 \ra \Real^3$ from pixel values $\Pixels$ imaged on the sensor. The scene is illuminated by a set of laser shots directed towards the visible geometry, hitting at positions $\Lights$; an image is captured for each shot. \FigFull{scene} shows an example of our setup, where $\light\in\Lights$ is a virtual point light source created by the laser's reflection, and $\vispoint\in\Visible$ is the projection of pixel $\pixel$ in the known geometry $\Visible$ as $\fProjection(\pixel)=\vispoint$, with inverse mapping $\fInvProjection(\vispoint)=\pixel$. It can be seen how the total propagation time from a hidden surface point $\px$ forms an ellipsoid with focal points at $\light$ and $\vispoint$.

In our particular context, the captured signal $\Signal$ is a time-resolved image indexed by the spatial and the temporal domains, measuring the light's time of arrival on each pixel. Physically, the signal formation model is
\begin{align}
\Signal(\pixel,\sT)  & = \int_\Scene \Rado(\px\ra\vispoint,\sT-\fOptDepth(\px\ra\vispoint)) \, \fGeom(\px \ra \vispoint) \fVis(\px \ra \vispoint) \diff \px, 
\label{eq:imageform1}
\end{align}
where $\Signal(\pixel,\sT)$ is the captured signal for pixel $\pixel$ at time instant $\sT$, $\Rado(\px\ra\vispoint, \sT-\fOptDepth(\px\ra\vispoint))$ is the reflected radiance at $\px$ towards $\vispoint$, and $\fOptDepth(\px\ra\vispoint))=\|\px-\vispoint\|\,\sSOL^{-1}$, with $\sSOL$ the speed of light in the medium. $\fGeom$ and $\fVis$ are respectively the geometric attenuation and binary visibility function between $\px$ and $\vispoint$.

Assuming that all recorded radiance at $\px$ is due to third bounce reflection, we can remove the visibility term $\fVis$. We also consider that all captured illumination comes from the virtual point light at $\light$. With that in place, we transform Eq.~\eqref{eq:imageform1} to
\begin{align}
\Signal_\light(\pixel,\sT)  & = \int_\Scene \Rado(\light\ra\px, \sT_o)\,\fBRDF(\light\ra\px\ra\vispoint)\,\fGeom(\light \ra \px \ra \vispoint) \diff \px, 
\label{eq:imageform2}
\end{align}
where  $\fBRDF$ represents the BRDF of the material at $\px$, $\fGeom(\light \ra \px \ra \vispoint) = \fGeom(\light \ra \px) \fGeom(\px \ra \vispoint)$ is the geometric attenuation of the full light path, and $\sT_o=\sT - \fOptDepth(\light\ra\px\ra\vispoint) - t_s - t_p$. Note that Eqs.~\eqref{eq:imageform1} and \eqref{eq:imageform2} assume that the time coordinate starts when $\light$ emits light, and that the sensor is placed at $\pixel$. $t_s$ and $t_p$ are the times of flight from the laser to the virtual light $\light$, and from  the camera to the imaged surface point $\vispoint$, respectively. However, since these additional terms do not affect the shape of the ellipsoids, we omit them in the rest of the paper for clarity (although we do take them into account in our calculations). 



\subsection{NLOS by back-projection} Taking into account the signal formation model~\eqref{eq:imageform2}, we would like to recover the unknown geometry $\Scene$ from a set of spatio-temporal measurements $\Signal_\light(\pixel,\sT)$. Back-projection methods~\cite{Velten2012nc,Buttafava2015} aim to recover a discrete approximation $\Voxels$ of the unknown scene $\Scene$. 
Intuitively, only points $\px\in\Scene$ where there is an object will scatter light back towards $\Visible$. However, since we only know the spatial and temporal domain of the signal, given a measurement $\Signal_\light(\pixel,\sT)$ there is an infinite number of candidate points $\px\in\Scene$ that might have reflected light towards $\fProjection(\pixel)=\vispoint$ at instant $\sT$ (see \Fig{scene}). This means that the scene cannot be recovered from a single measurement, so instead multiple measurements need to be taken. 
%

In essence, the main idea is to build a probabilistic model where, assuming Lambertian reflectances, the probability of point $\px$ being part of the occluded geometry is modeled as
\begin{align}
\pdf(\px)  & = \iiint \frac{\Signal_\light(\pixel,\sT) \, \delta(\sT-\fOptDepth(\light\ra\px\ra\vispoint))}{\fGeom(\light \ra \px \ra \vispoint)} \, \diff{\sT} \diff{\vispoint} \diff{\light}  \\
& \approx \sum_{\light \in \Lights}\sum_{\pixel \in \Visible} \Signal_\light(\pixel,\fOptDepth(\light\ra\px\ra\vispoint))\,\fGeom(\light \ra \px \ra \vispoint)^{-1},
\label{eq:probability_map}
\end{align}
where the signal $\Signal_\light(\pixel,\sT)$ is corrected by an estimate of $\fGeom(\light \ra \px \ra \vispoint)$, and $\delta$ is the delta function centered at zero. This probability map is later used to reconstruct the geometry, by using some operator over the voxelized representation, typically a three-dimensional Laplacian filter~\cite{Velten2012nc}.

In practice, the most common straight forward form of computing the probability map consists of evaluating Eq.~\eqref{eq:probability_map} for each unknown point $\px\in\Voxels$. This unfortunately is very expensive; the computational cost grows linearly with the number of pixels $\NumPixels$, lights $\NumLights$, and voxels $\NumVoxels$, thus yielding $\Order{\NumPixels\times\NumLights\times\NumVoxels}$.  In the following, we introduce our novel formulation, which significantly reduces the theoretical complexity of the computations, and allows for a very efficient implementation in commodity hardware.

\section{Our method}
\label{sec:our_method}

Our method builds on the observation that the set of points that can potentially contribute to $\Signal_\light(\pixel,\sT)$ from a given laser shot hitting at $\light$ is defined by the ellipsoid $\sElipsoid(\light,\vispoint,\sT)$ with focal points at $\light$ and $\vispoint$, and focal distance $\sT\cdot{\sSOL}$ (see \Fig{scene}). 
This means that the more ellipsoids intersecting at point $\px$, the higher the probability $\pdf(\px)$ of having occluded geometry at $\px$, since more light arriving at pixel $\pixel$ can have potentially been reflected at $\px$. 

Following this observation, we pose Eq.~\eqref{eq:probability_map} as an \textit{intersection of ellipsoids} $\sElipsoid(\light,\vispoint,\sT)$ with 
a voxelized representation of the scene, as 
%
\begin{align}
\pdf(\px)  = \sum_{\light \in \Lights}\sum_{\pixel \in \Visible} \Signal_\light(\pixel,\sT) \, \textrm{isect}(\px,\sElipsoid(\light,\vispoint,\sT)) ,
\label{eq:probability_map_ellipsoids}
\end{align}
where $\sT=\fOptDepth(\light\ra\px\ra\vispoint)$, and $\textrm{isect}(\px,\sElipsoid(\light,\vispoint,\sT))$ is a binary function returning 1 if the ellipsoid $\sElipsoid(\light,\pixel,\sT)$ intersects voxel $\px$, and 0 otherwise. 
Note that here we are not correcting for the geometric attenuation as in Eq.~\eqref{eq:probability_map}; we opt for this approach to avoid some possible numerical singularities when $\fGeom(\light \ra \px \ra \vispoint)\ra 0$, and to keep the maximum values bounded (which is important in our implementation to avoid overflow, see \Apx{imp_details}). Given that the probability map is latter processed by the Laplacian filter to reconstruct geometry, and the geometry term has almost no effect between neighboring voxels, we find that this does not affect the final result.

Our new formulation would be slower than Eq.~\eqref{eq:probability_map} if computed naively, due to the need to calculate the intersection between the ellipsoid and point $\px$. 
However, the intersection operand allows us to compute $\pdf(\px)$ by testing the ellipsoid-voxel intersection directly. Instead of evaluating each voxel $\px$ against the captured data, we now simply project the captured data into the voxelized scene representation. 

Posing the problem this way has two main benefits: On the one hand, it significantly reduces the complexity of the required computations by only testing on locations with signal information, resulting in a theoretical complexity order of $\Order{\NumPixels\times\NumLights\times\NumFrames}$, with $\NumFrames$ the temporal resolution. Note that this complexity is independent on the voxelization resolution, and that since the captured signal is in general sparse, in practice the complexity is even lower. 
%
On the other hand, our new formulation is equivalent to performing a voxelization of the full set of ellipsoids defined by the combination of tuples $\langle \vispoint, \light, \sT\rangle$ (\Fig{scene}, right). Voxelization is a well-studied problem in computer graphics, which can be efficiently performed in commodity hardware~\cite{Schwarz2010}. In the following we describe the details of our implementation.

\subsection{Fast back-projection}
\label{sec:gpu}
In order to perform the voxelization of the ellipsoids we rely on hardware-based voxelization~\cite{Eisemann2006}, although other custom GPU-based voxelization methods could be used (e.g.~\cite{Schwarz2010,Pantaleoni2011}). 
%

%
%
To achieve this, we need to overcome two main problems: \emph{i)} we need to create a large number of ellipsoids, which can be very expensive and memory consuming; and \emph{ii)} hardware rasterization does not work with parametric surfaces beyond triangles, so we need to tessellate the ellipsoids before rasterization; this aggravates the cost/memory problem. 

We address the first point by taking advantage of instanced rendering, which is standard in most modern GPUs, and allows to re-render the same primitive while applying a different linear transformation to each instance. For this, we create a base sphere, which is later transformed in the target ellipsoid $i$ by scaling, translating and rotating it, by using a standard linear sphere-to-ellipsoid transformation matrix $\sEllipsoidTMatrix_i$. 

Before rendering, we apply recursive geodesic tessellation to the base sphere.
Ideally, we would like all triangles' sizes to be smaller than the voxel size, so that high curvatures are accurately handled. However, since the transformations required for each ellipsoid might vary, the final size of each rendered triangle is not known in advance; this means that we cannot set a particular tessellation level for all ellipsoids. We instead precompute a set of spheres $\sTessSpheres$ with different tessellation levels $\sTessLevel$, and dynamically choose what level $\sTessLevel$ will be used as
%
\begin{align}
 \textrm{argmin}_{\sTessLevel \in \sTessSpheres} \, \left(\AreaTriangle_\sTessLevel \max( \textrm{Eig}(\sEllipsoidTMatrix_i)) < \epsilon\right),
\label{eq:sphere_selection}
\end{align}
where $\AreaTriangle_\sTessLevel$ is the approximation error per triangle at tessellation level $\sTessLevel$, $\textrm{Eig}(\sEllipsoidTMatrix_i)$ are the eigenvalues of transformation matrix $\sEllipsoidTMatrix_i$, and $\epsilon$ is an error threshold, which we set to the voxel size.  We offer additional implementation details in the Appendix. 

\subsection{Cost analysis}
\label{sec:cost_order}
The computational cost of our method is linear with the number of ellipsoids $\Order{\NumPixels\times\NumLights\times\NumFrames}$, which is independent on the resolution of the reconstruction space. Voxelizing each ellipsoid has a linear cost with the number of triangles per ellipsoid $\NumTriangles$ times the cost of each triangle $\bigtriangleup$, which is approximately proportional to the number of voxels  $\NumVoxels_\bigtriangleup$ containing the triangle. Thus, the cost for each ellipsoid $\Order{\sElipsoid}$ is:
\begin{equation}
\Order{\sElipsoid} = \twopartdef{\Order{\NumTriangles\times\NumVoxels_\bigtriangleup}\approx\Order{\sqrt[3]{\NumVoxels}}}{\NumVoxels_\bigtriangleup \geq 1}{\Order{\NumTriangles}}.
\end{equation}
As we will show later, setting  $\epsilon$ [Eq.~\eqref{eq:sphere_selection}] so that $\NumVoxels_\bigtriangleup \approx 1$ gives the best trade-off between quality and cost. With this cost per ellipsoid, the total order of our technique is  $\Order{\NumPixels\times\NumLights\times\NumFrames\times\sqrt[3]{\NumVoxels}}$, which results in a speed-up with respect to traditional back-projection of $\Order{\frac{\sqrt[3]{\NumVoxels^2}}{\NumFrames}}$. In the following section we demonstrate empirically the performance of our method.

\section{Results}
\label{sec:results}

We evaluate our method by using datasets from three different sources (captured with femto-photography, a SPAD, and generated with transient rendering), each showing variable ranges of complexity, as well as different levels of signal quality. We compare our method against traditional back-projection~\shortcite{Velten2012nc}, using Velten et al.'s optimized implementation. All our tests have been performed on an Intel i5-6500 @3.2GHz with 8 GB of RAM equipped with a GPU Nvidia GTX 1060.

\begin{figure}[t]
  \centering
  \def\svgwidth{1\textwidth}\footnotesize
  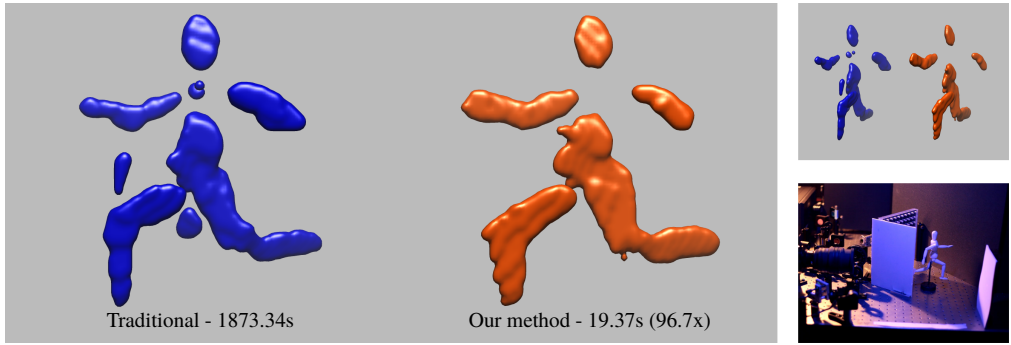
  \caption{Reconstruction of a mannequin (see bottom right) captured with a streak camera and a femtosecond laser~\cite{Velten2013}, reconstructed using traditional back-projection (left, in blue) and our method (right, in orange), which is two orders of magnitude faster while yielding similar quality.  The inset on the top-right shows the same reconstructed object under a different camera angle (inset from \cite{Velten2012nc}).  
}
  \label{fig:gandalfUp}
\end{figure}

\FigFull{gandalfUp} shows the reconstruction of a hidden mannequin captured using a streak camera and a femtosecond laser~\cite{Velten2013}.
A voxel grid of resolution $\NumVoxels=162^3$ is reconstructed from a set of $\NumLights=59$ spatio-temporal measurements with different lighting position $\light$. Each measurement has spatial resolution of $\NumPixels=336$ pixels, and temporal resolution of  $\NumFrames=512$ frames.  
Although our method is mathematically equivalent to traditional back-projection (see \Sec{our_method}), reconstructing this scene takes less than 20 seconds with our method, compared to more than half an hour with traditional back-projection. This represents a speed-up of 96.7x. 

\begin{figure}[t]
  \centering  
    \def\svgwidth{.9\textwidth}\footnotesize
    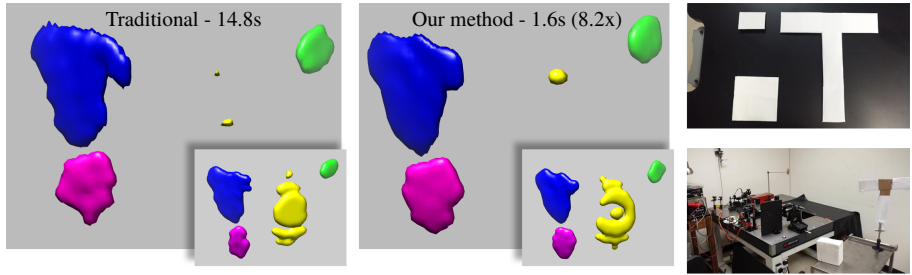
  \caption{Reconstruction of the scene captured using a SPAD~\cite{Buttafava2015} using traditional back-projection (left), and our method (right). The rightmost images show the capture setup (insets from~\cite{Buttafava2015}).
We have used a different color to differentiate each object (blue: T; pink: big patch; green: small patch; yellow: noise from the camera). 
The quality of the reconstruction is almost identical. Note that the original dataset had a higher amount of camera noise, which we have filtered before reconstruction (the insets show the reconstruction with each method for the original unfiltered data). Our method takes only 1.8s to reconstruct the scene, in comparison to 14.8s for traditional back-projection. 
}
  \label{fig:spad}
\end{figure}

\FigFull{spad} shows a similar reconstruction using recent data obtained using a single-photon avalanche diode (SPAD)~\cite{Buttafava2015}. The SPAD is a more affordable transient imaging device, at the cost of lower quality measurements (i.e. less spatio-temporal resolution and higher levels of noise). 
A voxel grid of resolution $\NumVoxels=150^3$ is reconstructed from a set of $\NumLights=185$ temporal measurements with different lighting position $\light$. Each measurement has a single pixel $\NumPixels=1$, and a temporal resolution of $\NumFrames=14000$ frames.
Note that this is a difficult scenario for our method, given the high levels of noise in the captured signal, as well as the large temporal resolution; however, our method takes only 1.8s to reconstruct the scene, in comparison to 14.8s for traditional back-projection.

\subsection{Analysis}
\newcommand{\fithtextwidth}{.14\textwidth}
\bgroup
\setlength{\tabcolsep}{2pt}
\renewcommand{\arraystretch}{1}
\begin{figure}
\centering
\begin{tabular}{cccccc}
&$\NumVoxels=16^3$&$\NumVoxels=32^3$&$\NumVoxels=64^3$&$\NumVoxels=128^3$&$\NumVoxels=256^3$\\ 
\raisebox{.4cm}{\rotatebox[origin=l]{90}{Ours}} 	&\includegraphics[width=\fithtextwidth]{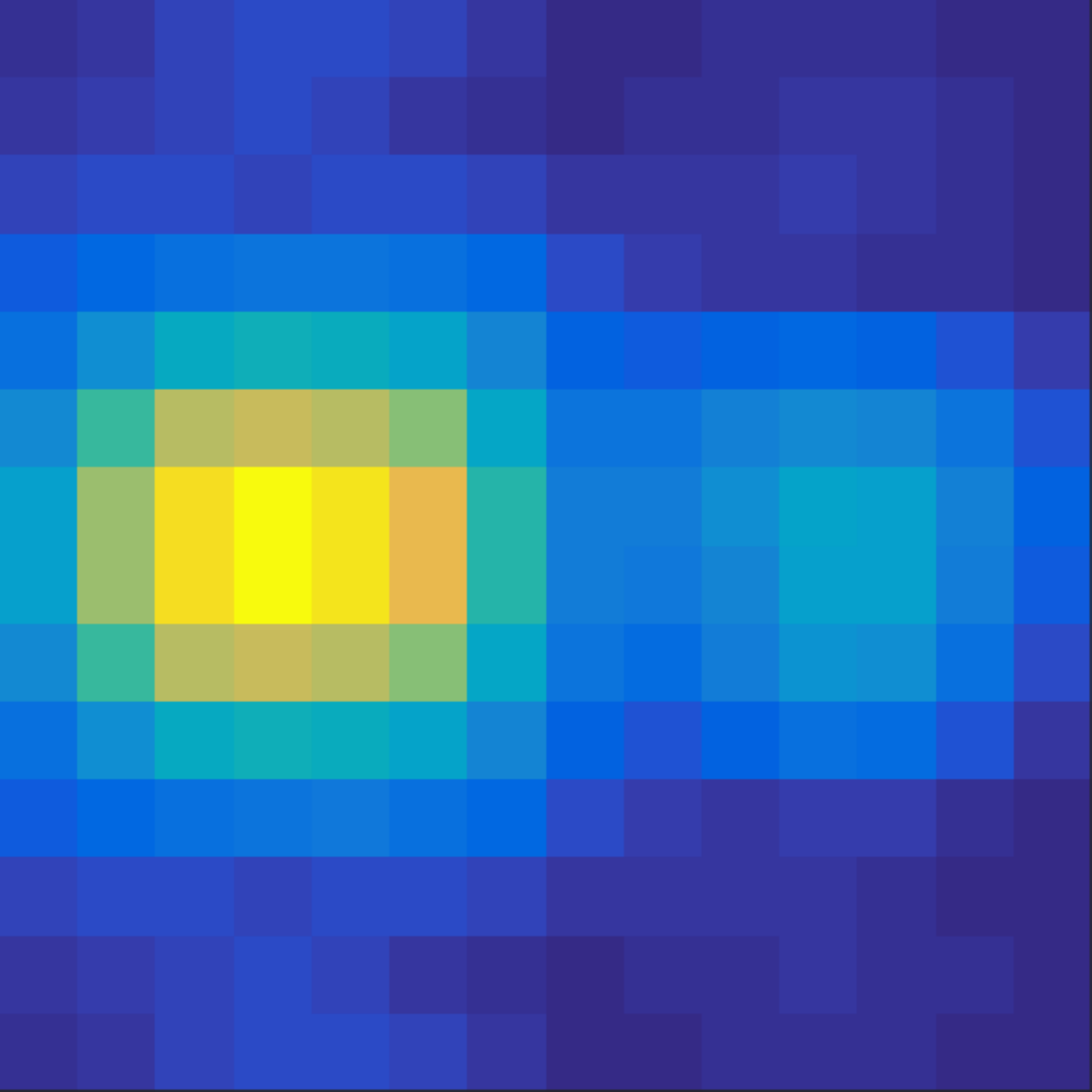}&
	\includegraphics[width=\fithtextwidth]{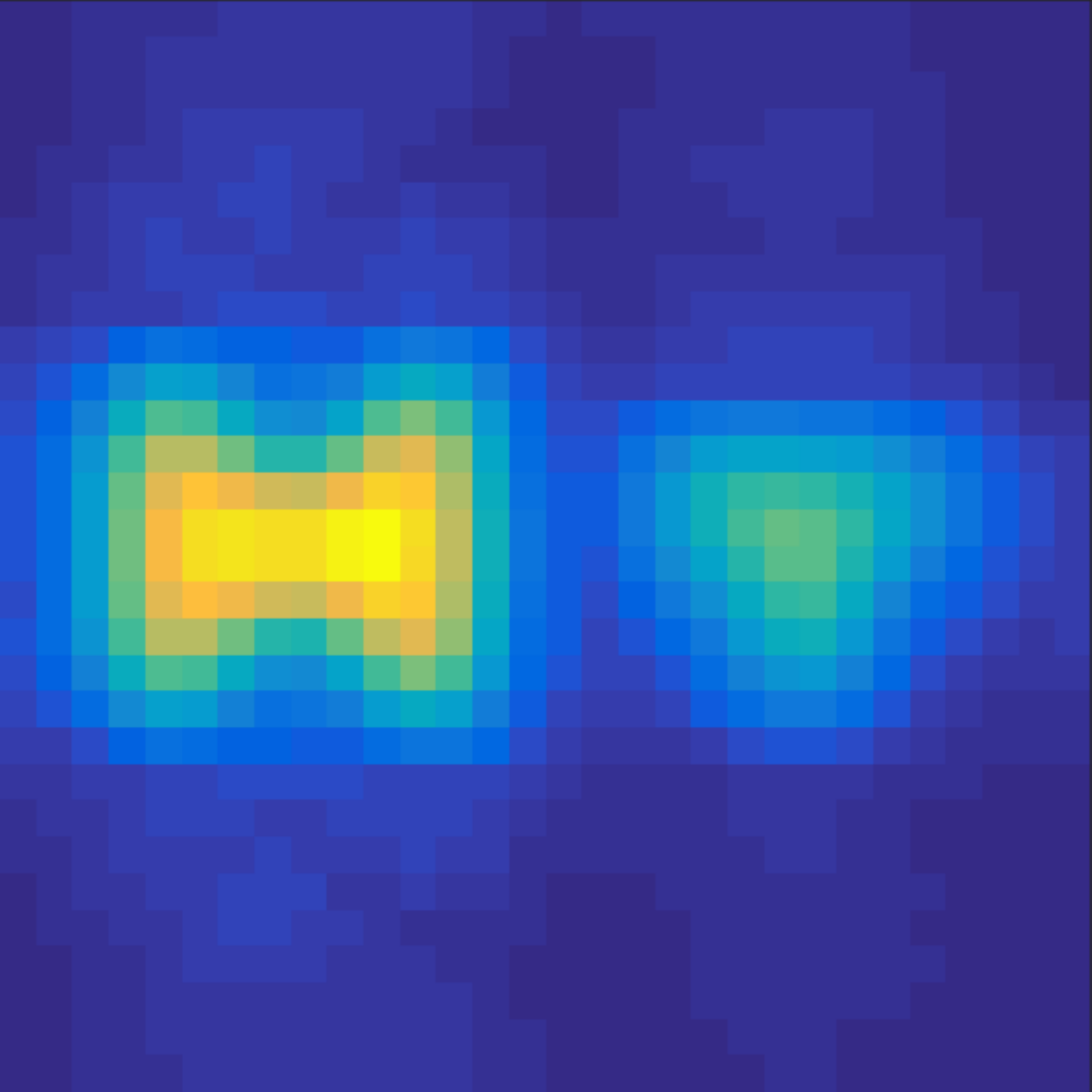}&
  	\includegraphics[width=\fithtextwidth]{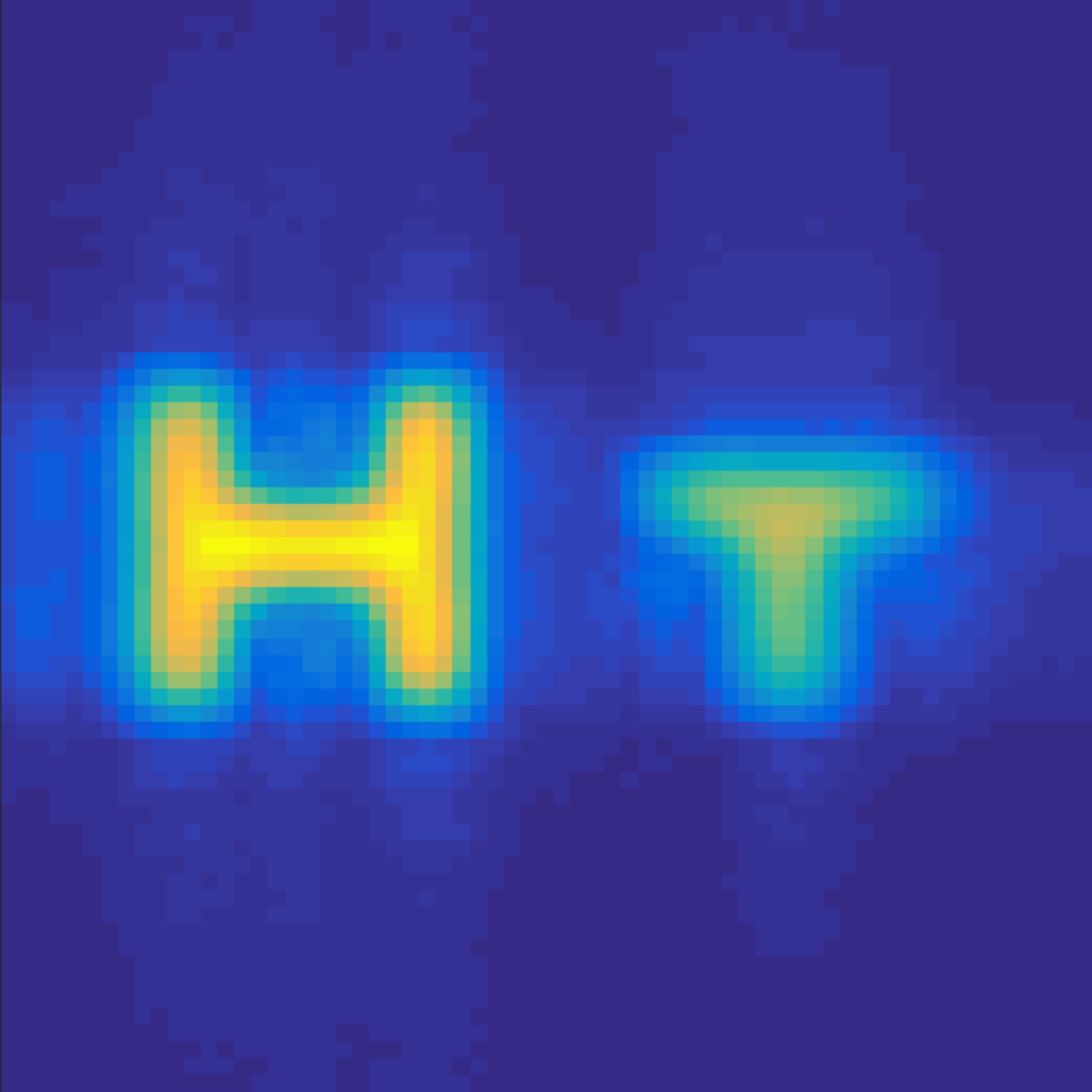}&
  	\includegraphics[width=\fithtextwidth]{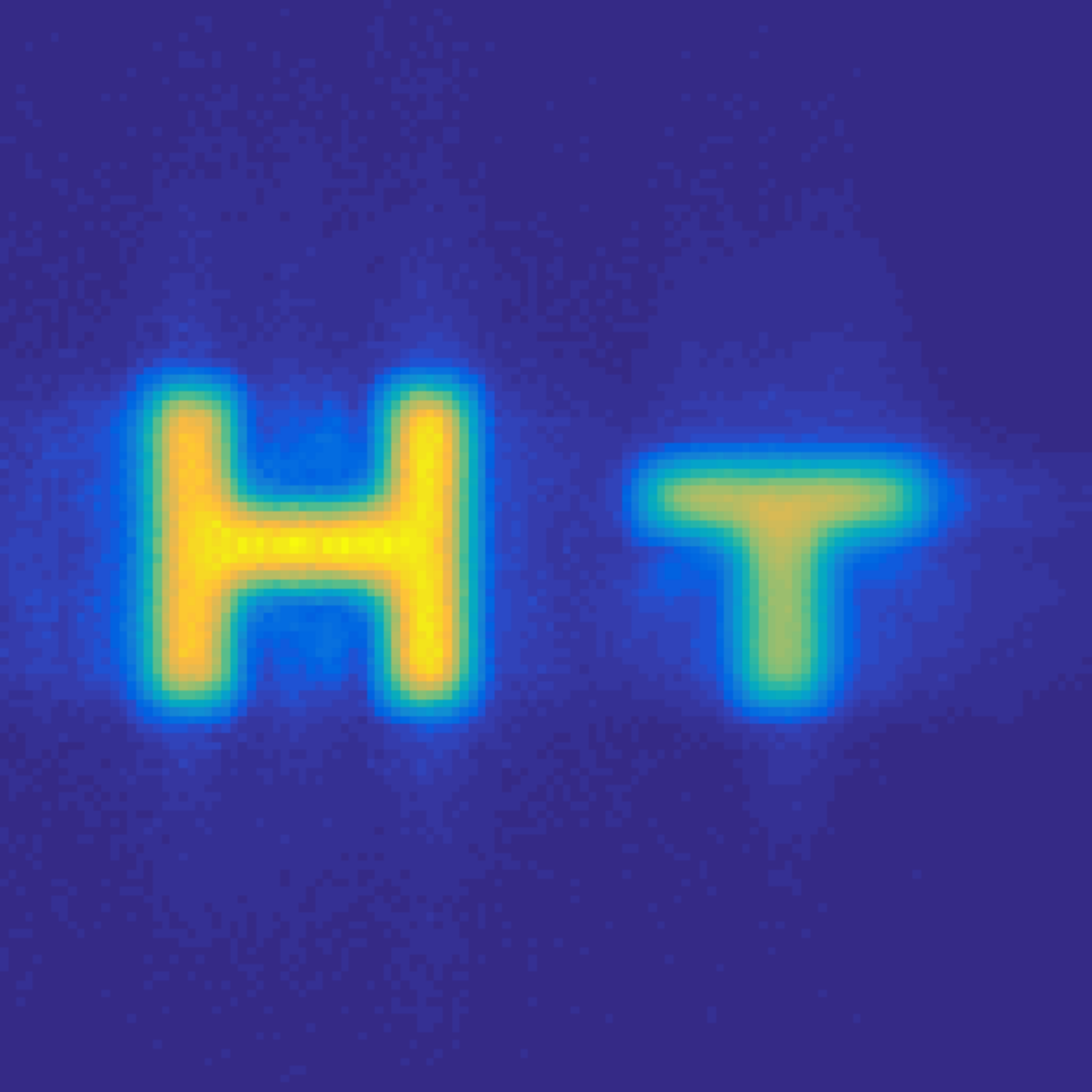}&
  	\includegraphics[width=\fithtextwidth]{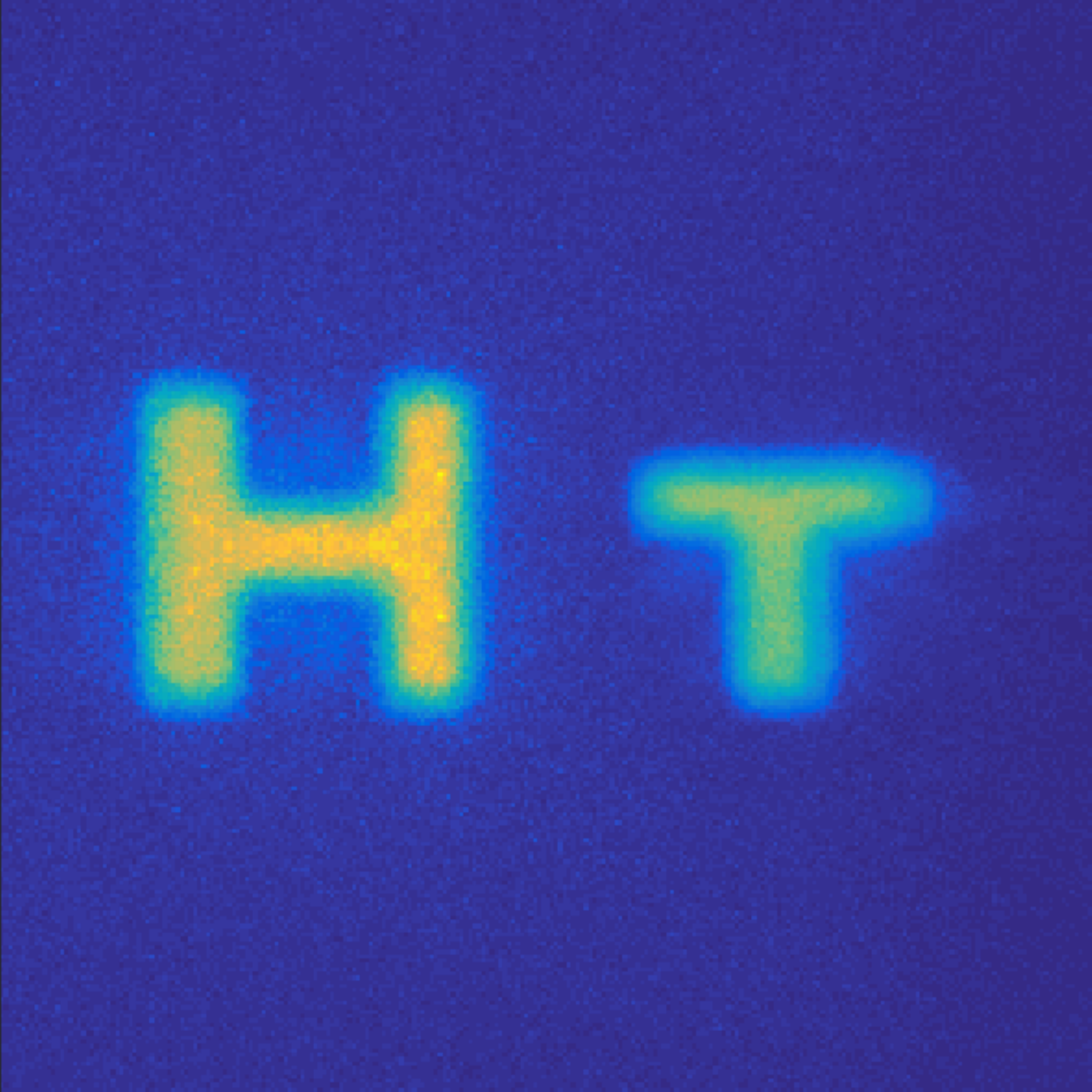}
\\
\raisebox{-.1cm}{\rotatebox[origin=l]{90}{Traditional}} &	\includegraphics[width=\fithtextwidth]{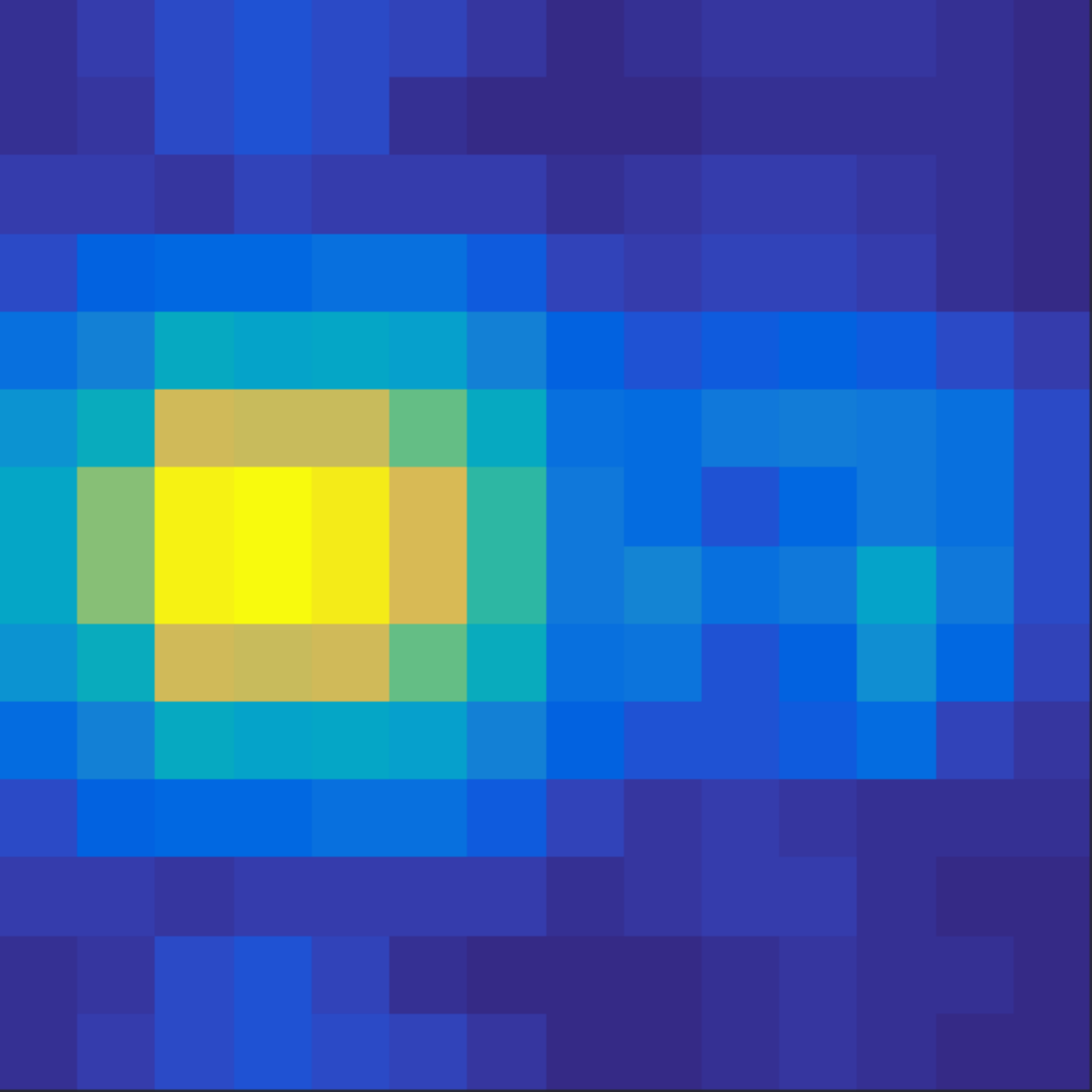}&
  	\includegraphics[width=\fithtextwidth]{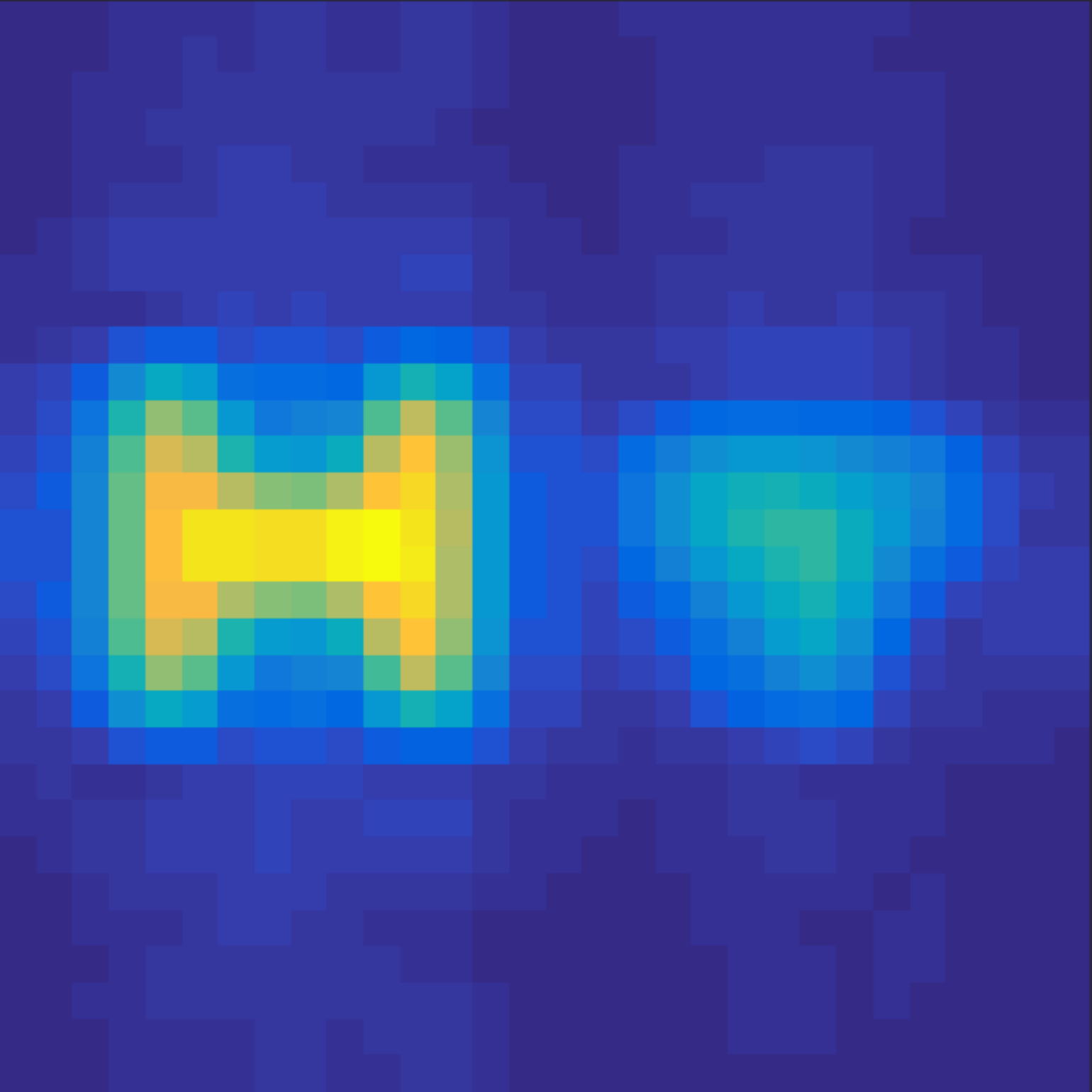}&
  	\includegraphics[width=\fithtextwidth]{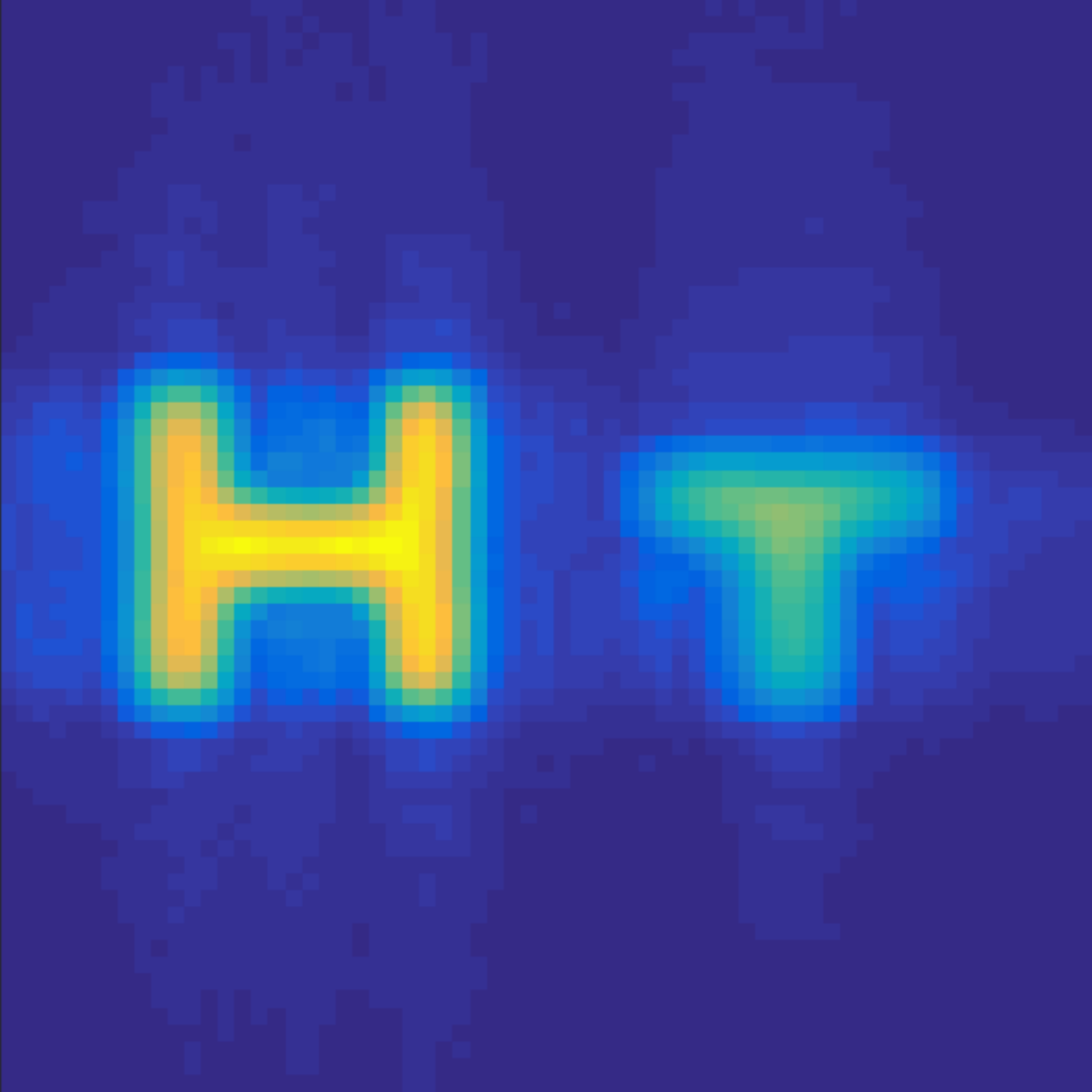}&
  	\includegraphics[width=\fithtextwidth]{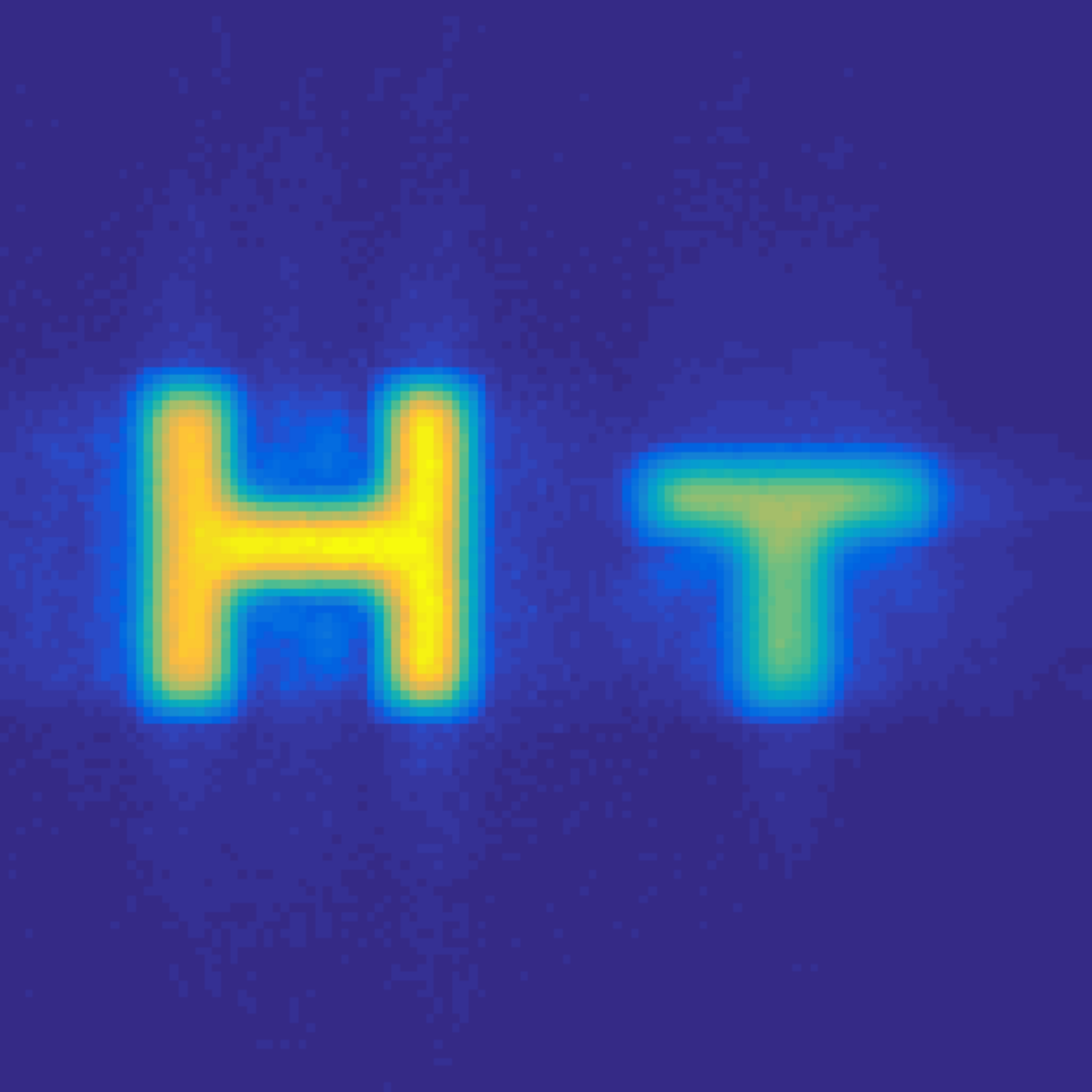}&
  	\includegraphics[width=\fithtextwidth]{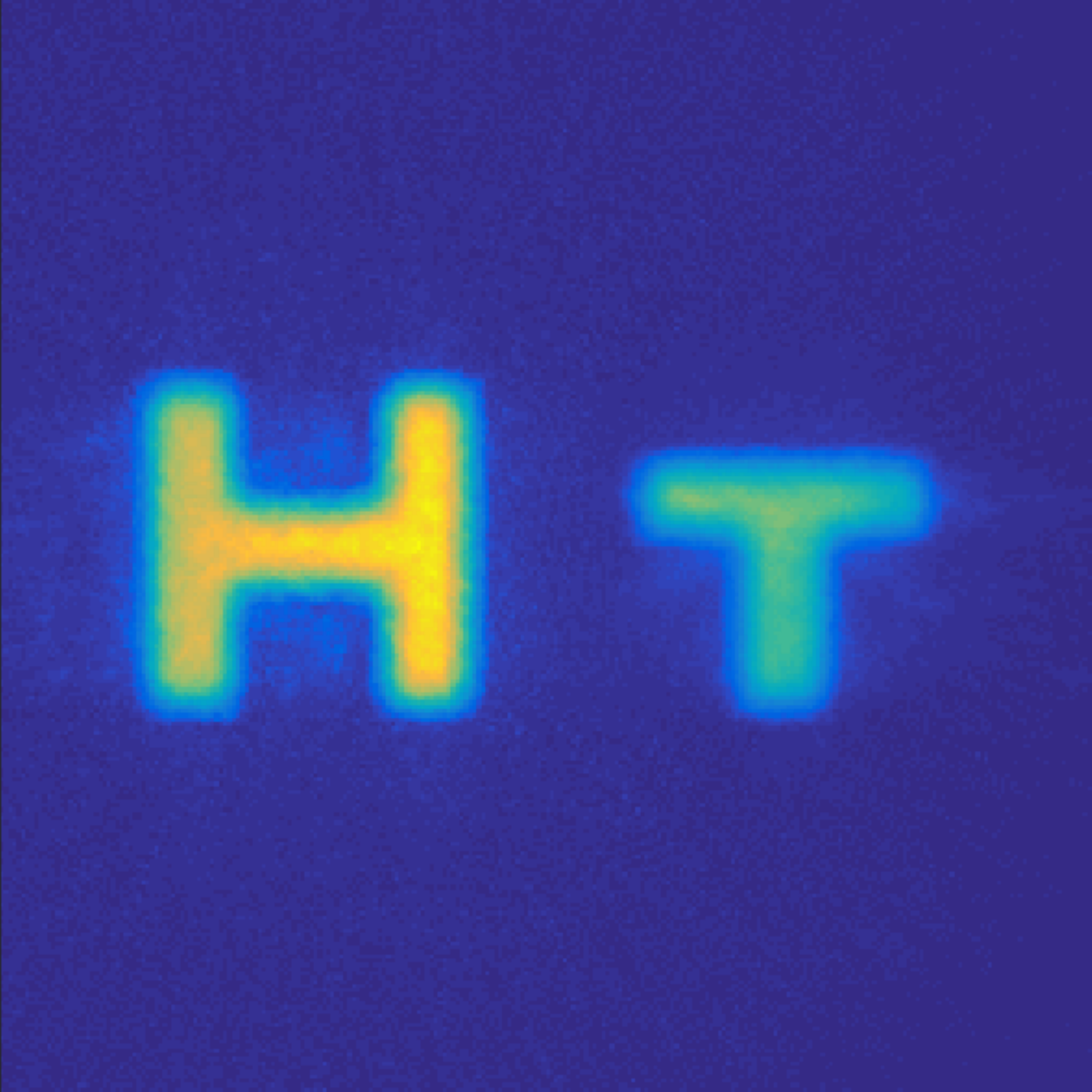}
\end{tabular}
\caption{Comparison between the reconstruction quality computed with our method (top) and traditional back-projection (bottom), for an increasing number of voxels $\NumVoxels$. \new{A comparison of the cost of both algorithms, as well as the progression of their numerical error can be found in \Fig{sizesBenchmarks}, left.}}
%
\label{fig:reconstructionComparisonDimensions}
\end{figure}
\egroup
We compare the performance of both our method and traditional back-projection~\cite{Velten2012nc} varying the three parameters involved in the reconstruction: The resolution of the output voxelized space $\NumVoxels$, the input spatial and temporal resolution $\NumPixels$, and $\NumFrames$. We use a synthetic scene (\Fig{reconstructionComparisonDimensions}) using the time-resolved rendering framework by Jarabo et al.~\shortcite{Jarabo2014}. The scene is similar to scenes captured in previous works~\cite{Heide2014mirrors,Buttafava2015}. We opt for a synthetic scene to eliminate errors that depend on the camera and the capture setup, and obtain a clean ground truth solution. 

\begin{figure}[!ht]
  \centering
  	\includegraphics[width=1\textwidth]{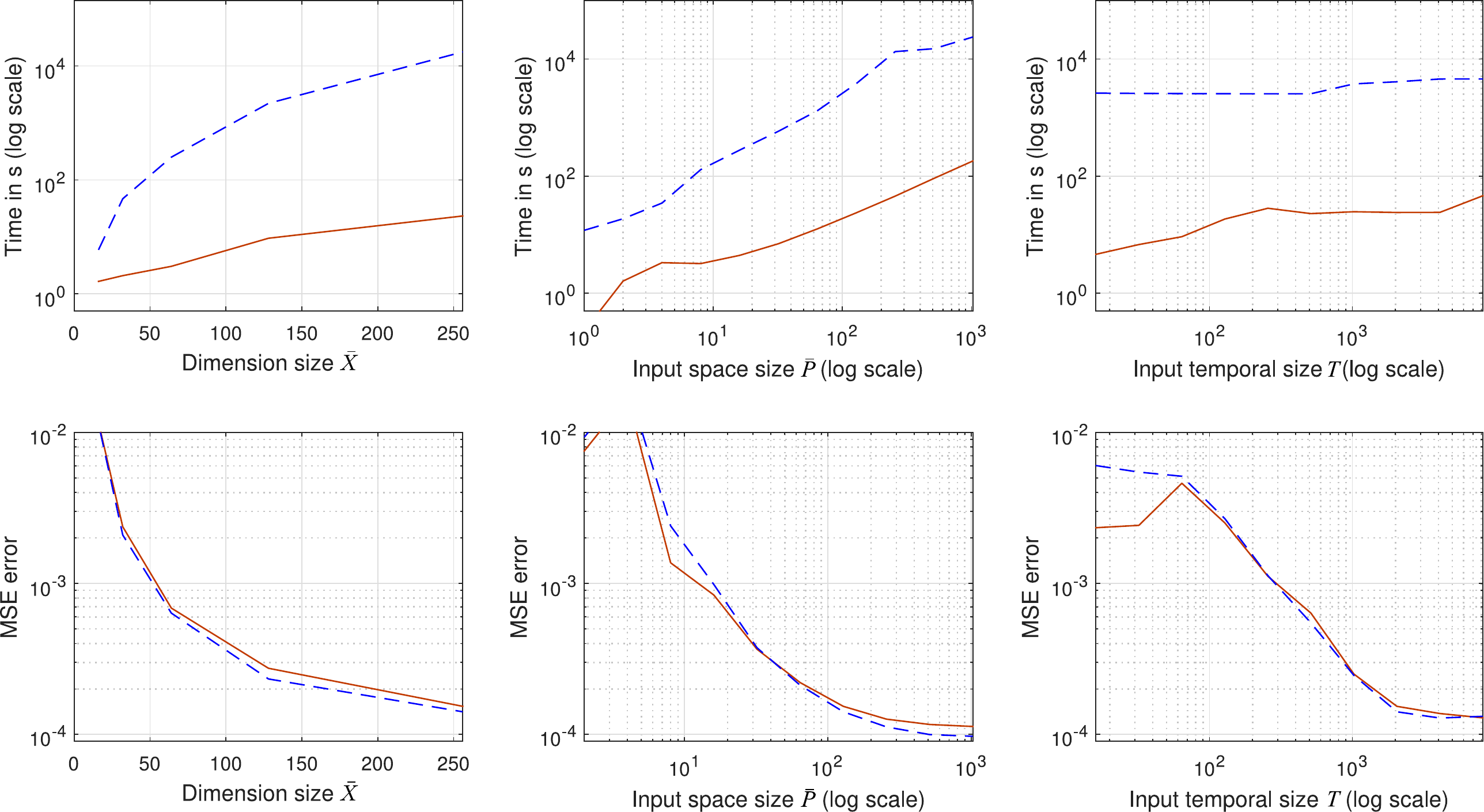}
  \caption{Cost (top) \new{and error with respect to the ground truth (bottom)} comparisons between our method (solid orange) and traditional back-projection (dashed blue) for the synthetic scene shown in \Fig{reconstructionComparisonDimensions}. Each graph varies one reconstruction parameters while fixing the other two, namely the number of voxels $\NumVoxels$ (right), input spatial resolution $\NumPixels$ (middle) and input temporal resolution $\NumFrames$ (right). We fix the parameters to a reconstruction space of $\NumVoxels=256^3$ voxels, an input spatial resolution $\NumPixels=128$ pixels, and temporal resolution $\NumFrames=1024$ frames. For all reconstruction we use 128 measurements with different virtual light positions $\light$. \new{As shown in the bottom, the error introduced by our algorithm with respect to traditional back-projection is negligible. Note that at low spatial $\NumPixels$ and temporal $\NumFrames$ resolutions the error in both cases is large and scene dependent, which is the reason for the counter-intuitive behavior at very low $\NumPixels$ (bottom center) and $\NumFrames$ (bottom right). }
  }
  \label{fig:sizesBenchmarks}
\end{figure}
\FigFull{sizesBenchmarks} \new{(top)} shows a comparison of the cost of traditional back-projection and our method, varying one parameter and fixing the other two. In all cases our method is significantly more efficient for all practical resolutions. In particular, our method shows a much better scalability than the previous work with respect to the number of reconstructed voxels $\NumVoxels$, showing speed-ups of up to 10000x for large  resolutions, while having similar convergence with respect to the number of input pixels $\NumPixels$. On the other hand, our method scales linearly with the number of frames $\NumFrames$, while the cost of traditional back-projection remains constant with respect to this parameter; however, note that even for large temporal resolutions (e.g. $10^4$ frames) our method is still two orders of magnitude faster. 
%
\new{In terms of reconstruction error, our algorithm scales similarly to traditional back-projection. \FigFull{reconstructionComparisonDimensions} shows the reconstruction result for a varying reconstruction resolution $\NumVoxels$, while \new{\Fig{sizesBenchmarks}  (bottom)} shows the error with respect to the ground truth for varying input parameters.} Thus, despite the cost is significantly lower with our algorithm, the error introduced is always comparable with the previous work.

\begin{figure}\color{white}\renewcommand\color[2][]{}
  \centering
	\def\svgwidth{0.24\textwidth}
    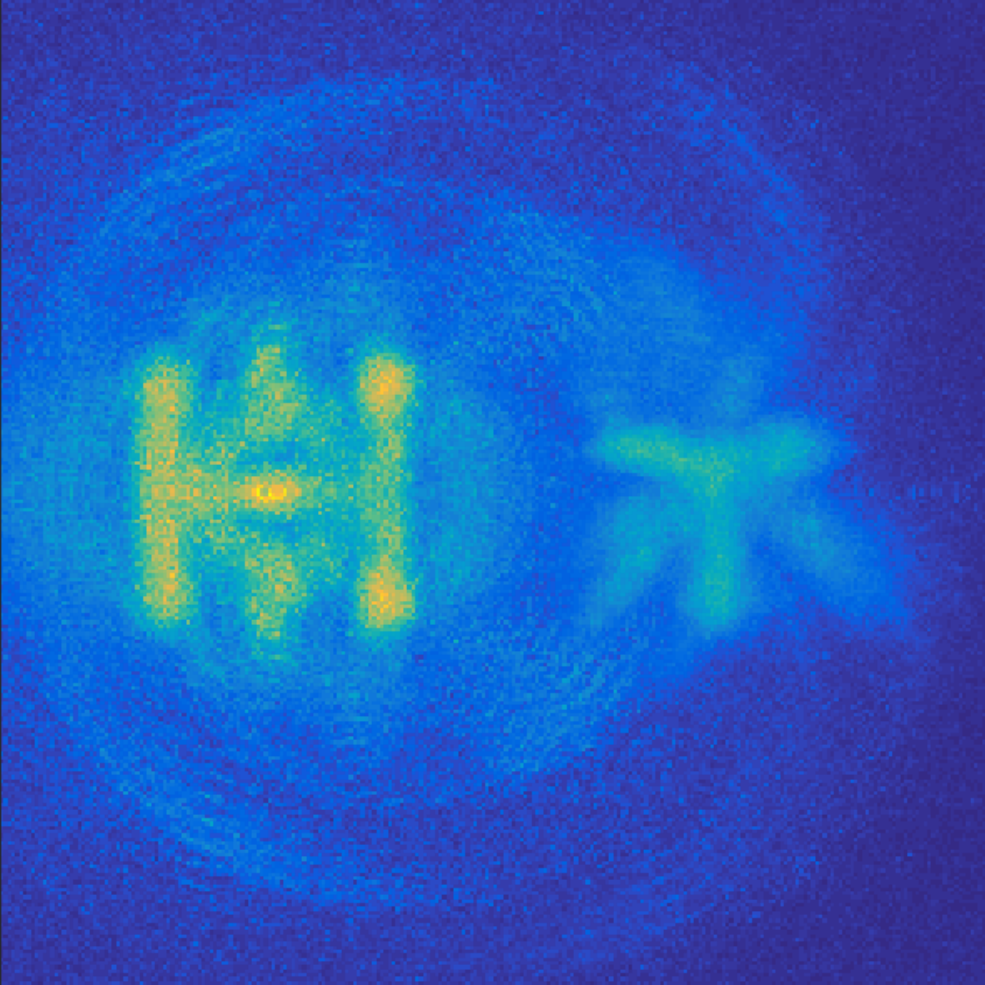
    \def\svgwidth{0.24\textwidth}
    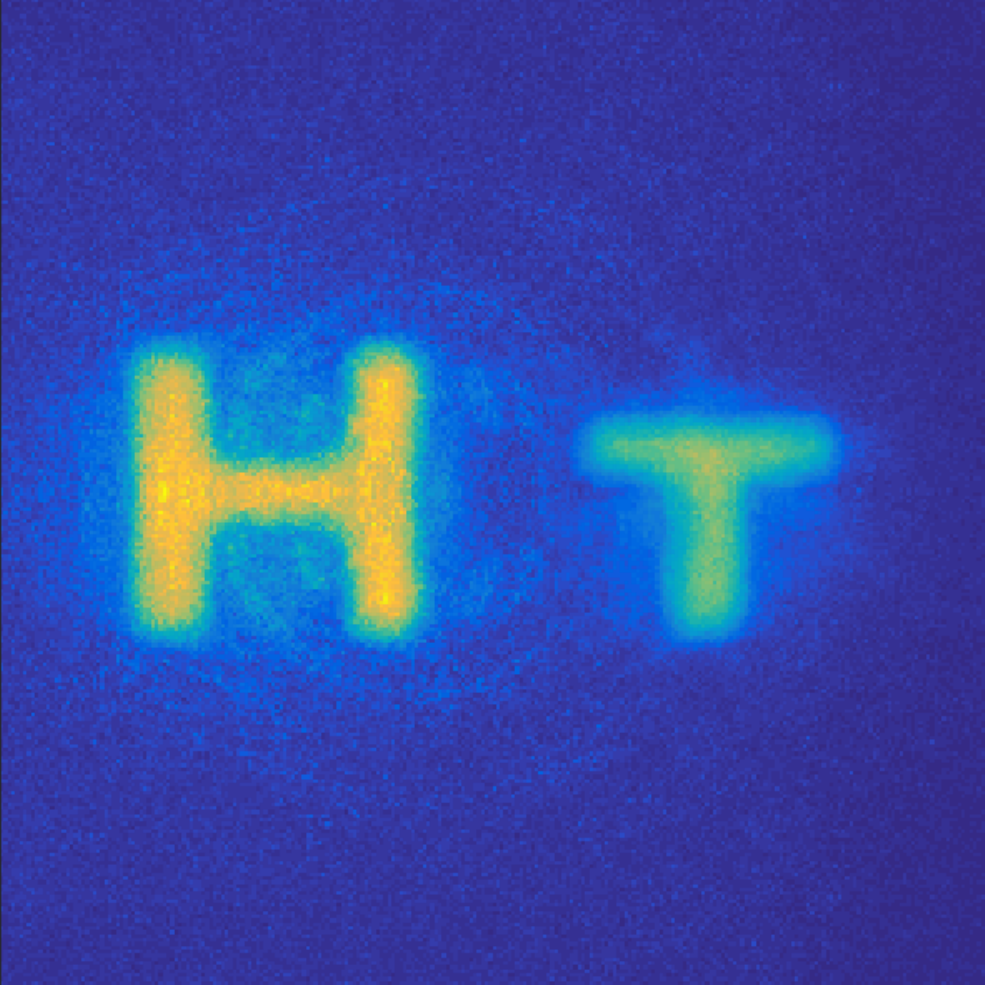
    \def\svgwidth{0.24\textwidth}
    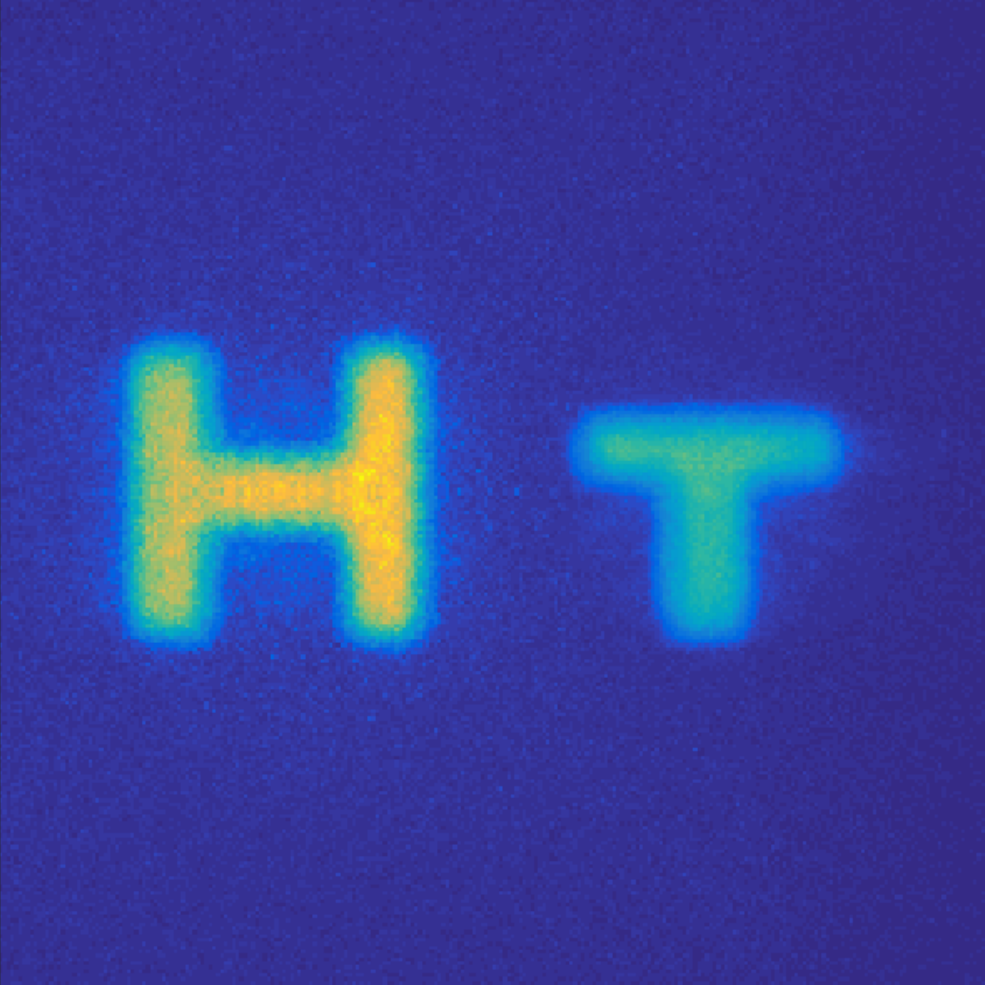
    \def\svgwidth{0.24\textwidth}
    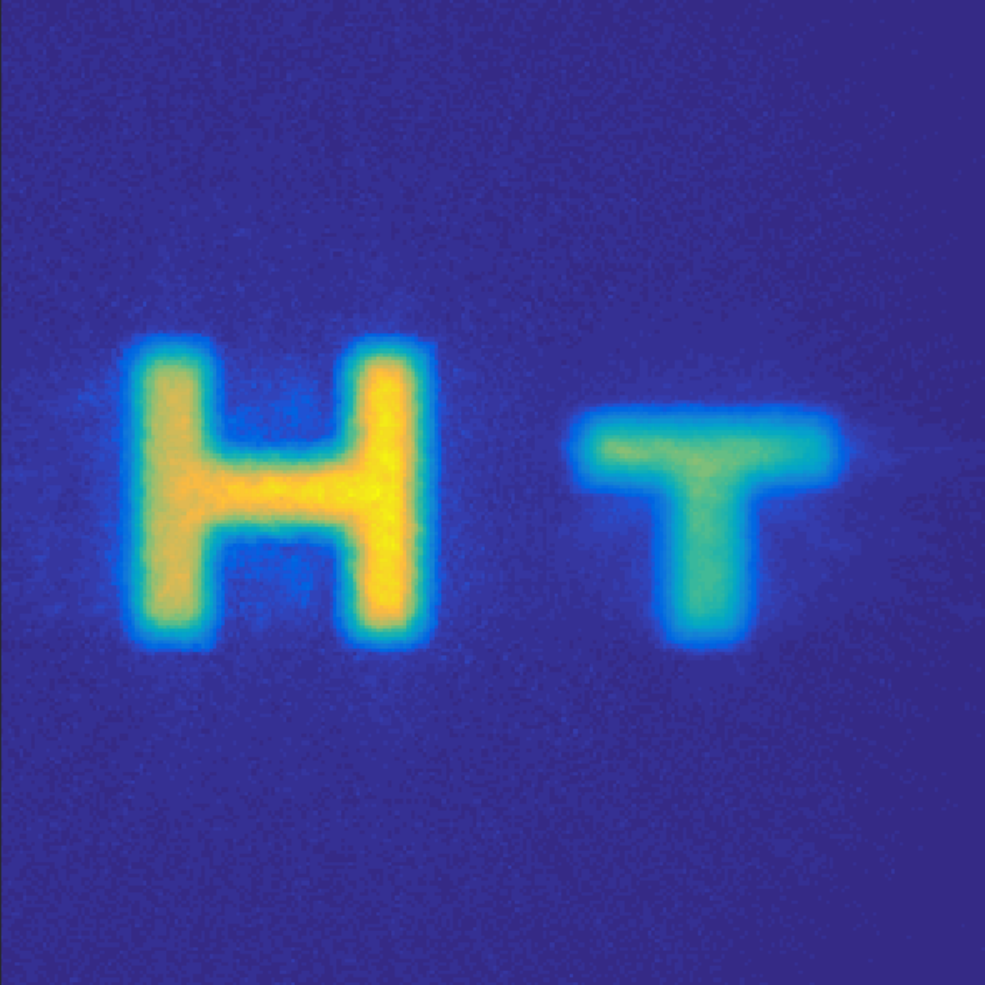
  \caption{Reconstruction results of our method with increasing ellipsoid tessellation quality (168, 656 and 2592 triangles per ellipsoid, respectively). The rightmost image shows the reconstruction result using using traditional back-projection. Our final reconstruction is comparable to traditional back-projection, computed with a speed-up of 1300x. }
\label{fig:reconstructionPrecisions}
\end{figure}

\begin{figure}[t]
	\centering
	\includegraphics[width=.8\textwidth]{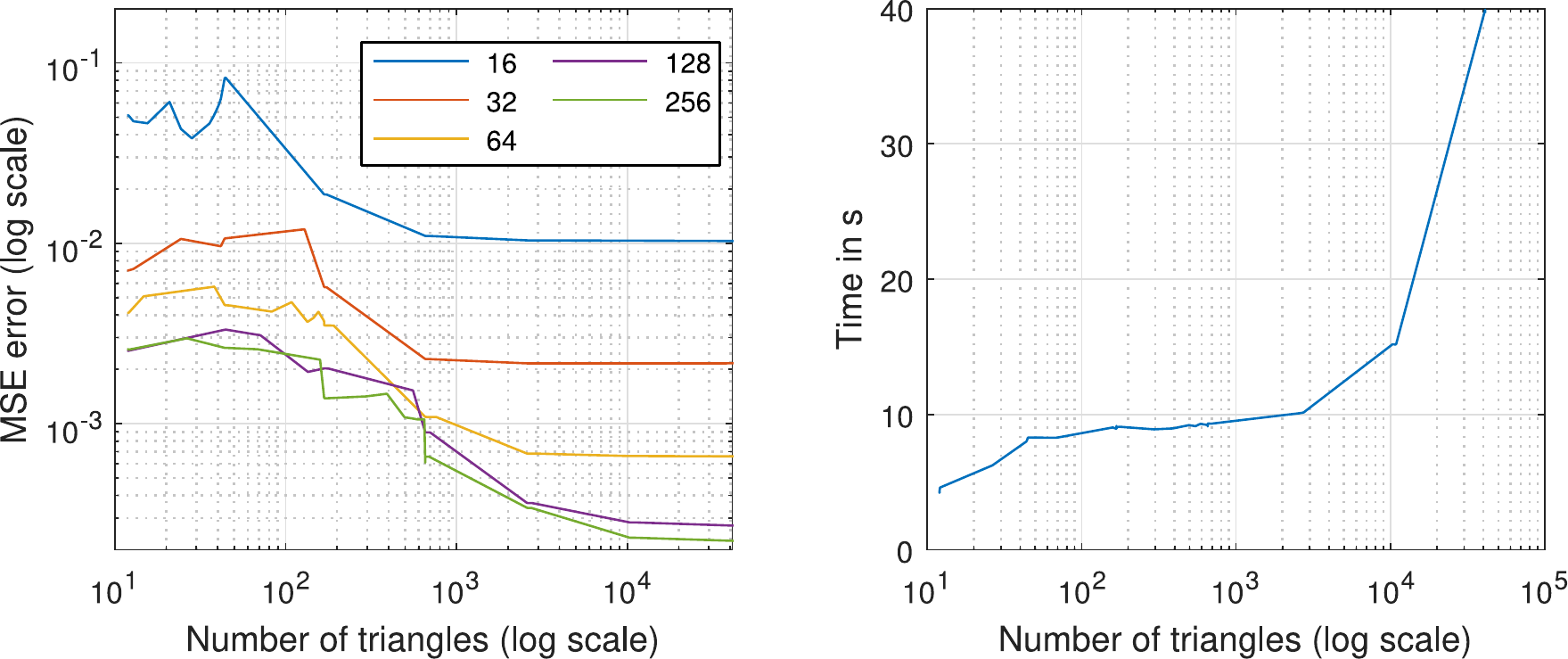}
  \caption{Left: MSE of our method with respect to a ground truth reconstruction for different voxel resolutions $\NumVoxels$, and for increasing number of triangles per ellipsoid. The minimum error is bounded by the reconstruction resolution, while increasing the number of triangles past a certain point does not improve the result. Right: Reconstruction time as a function of the number of triangles per ellipsoid used (reconstruction resolution of $\NumVoxels=256^3$). In practice we found that a near-optimal result can be obtained with a tessellation level of roughly the size of the voxel. }
  \label{fig:errorsCosts}
\end{figure}

An important additional parameter of our method is the quality of the ellipsoids' tessellation [Eq.~\eqref{eq:sphere_selection}]. This determines the number of triangles used to represent the ellipsoid during the voxelization, and has an impact in both the cost and the quality of the reconstruction. 
\FigFull{reconstructionPrecisions} shows the result with varying number of triangles, compared with the traditional method for a reconstructed space of $\NumVoxels=256^3$ voxels. For tessellation levels leading to triangles of approximately the voxel size, our reconstruction achieves a reconstruction quality similar to traditional back-projection, more than a thousand times faster. This is further illustrated in \Fig{errorsCosts}: We can see that our reconstruction quality is bounded by the reconstruction resolution, and at certain point increasing the number of triangles does not improve the result, while the cost increases linearly. In our tests we found that setting the triangle size to roughly the voxel size leads to a sweet-spot in terms of reconstruction quality and cost. 

\subsection{Discussion and limitations}
Our method takes advantage of the relatively sparse signal of time-resolved data to scale below the theoretic computational order (\Sec{cost_order}), by ignoring ellipsoids $\sElipsoid(\light,\vispoint,\sT)$ with $\Signal_\light(\pixel,\sT) \approx 0$. Thus, our worst-case scenario occurs for non-sparse signals. While this is not problematic in surface-based NLOS reconstruction, it becomes dominant in the presence of participating media. In these cases, our technique still performs significantly faster than traditional back-projection: For an input resolution of $\NumPixels=128^2$ and $\NumFrames=1024$ and a voxel resolution of $\NumVoxels=256^3$, traditional back-projection takes 475.78s, as opposed to 12.51s for for our method in the worst-case scenario. 

Our algorithm shares some of the limitations from previous back-projection-based NLOS methods. In particular, our probabilistic model [Eq.~\eqref{eq:probability_map}] assumes Lambertian surfaces, and ignores the effect of albedo.  Moreover, Eq.~\eqref{eq:probability_map} assumes that all light arriving the sensor is due to third-bounce reflection. While given the relative intensity between bounces is a sensible assumption, this might incur into errors in areas where higher-order scattering is dominant such as concavities. 

In addition, while our technique scales very well with respect the reconstruction resolution, the maximum reconstruction resolution is limited by the number of ellipsoids that can be generated from the input. If the input resolution is too low for the reconstruction voxelization, then several gaps due to insufficient ellipsoid coverage might appear. This can be solved by simply upsampling the input signal so that more ellipsoids can be generated, therefore increasing the cost in our algorithm. In this case the result would be similar to upsampling a lower-resolution reconstructed voxelized space. This imposes a maximum boundary on the resolution that can be achieved from a given input capture resolution.


\section{Conclusions}
\label{sec:conclusion}

Current reconstruction methods for NLOS reconstruction are still too slow for practical perfomance, becoming a key bottleneck of the process. In this work, we have presented a new method for NLOS reconstruction based on back-projection. 
Our work builds on the observation that the light incoming at a given pixel at instant $t$ can have been reflected only from the points defined by an ellipsoid with poles at the observed point and the light source, allowing to define the probability of the NLOS geometry as the intersection of several of these ellipsoids, and exploits it by posing NLOS reconstruction as a voxelization problem. 
This allows us to reduce the  computational cost significantly, and achieve speed-ups up to three orders of magnitude. Our technique can be efficiently implemented  on the GPU. We hope our work will help current and future NLOS reconstruction techniques, enabling their use in practical real-world setups: With that aim, we have made our code publicly available.

\appendix


\section{Implementation details}
\label{ap:imp_details}

We have implemented our voxelization in OpenGL 4.4. We set the frame buffer to a 3D single channel \texttt{UINT32} texture, with depth testing disabled since we want \textit{all} ellipsoids to be rendered. Most operations during rendering are performed in the geometry and fragment shaders: the former is used to select the best possible projection for tessellation (normal swizzling)~\cite{Schwarz2010}, which is important to avoid possible holes. In the latter we perform the drawing by means of an atomic add, which updates the voxel's probability $\pdf(\px)$. 

Due to limitations on current GPU architectures, we must set $\pdf(\px)$ as a \texttt{UINT32} value. To take this into account, we normalize the intensity values $\Signal_\light(\pixel,\sT)$ to 255. We set this maximum value to prevent data overflow, which would be reached when intersecting more than $2^{24}$ ellipsoids in a worst-case scenario. Note that this assumes that the signal's intensity can be represented within this range. If we would like to compute different bounces, they should be computed separated, to accommodate for the large differences in intensity between them. 

Although normal swizzling provides a good projection quality with almost no holes, there may still be some gaps in the probability map, due to fragment rejection when the triangle does not include the center of the projected pixel. While this could be fixed by conservative voxelization~\cite{Zhang2007}, 
multiple rasterizations of the same triangle may result in an overestimation of the probability map, so we choose not to apply it. 

Finally, writing data in GPU using arbitrary access to graphics memory does not guarantee coherency,
so we manage internal coherency manually by explicitly waiting for termination of write operations before filtering or rendering additional batches. 
%

\section*{Funding}
Defense Advanced Research Projects Agency - DARPA (HR0011-16-C-0025); European Research Council (ERC Consolidator Grant 682080); Spanish Ministerio de Econom\'{i}a y Competitividad (TIN2016-78753-P, TIN2014-61696-EXP).

\section*{Acknowledgements}
We want to thank Andreas Velten for providing the captured data, as well as the reconstruction code from~\cite{Velten2012nc}, and Julio Marco for his help setting up the figures.

\end{document}

%% file: setup.pdf_tex
\begingroup%
  \makeatletter%
  \providecommand\color[2][]{%
    \errmessage{(Inkscape) Color is used for the text in Inkscape, but the package 'color.sty' is not loaded}%
    \renewcommand\color[2][]{}%
  }%
  \providecommand\transparent[1]{%
    \errmessage{(Inkscape) Transparency is used (non-zero) for the text in Inkscape, but the package 'transparent.sty' is not loaded}%
    \renewcommand\transparent[1]{}%
  }%
  \providecommand\rotatebox[2]{#2}%
  \ifx\svgwidth\undefined%
    \setlength{\unitlength}{485.66274568bp}%
    \ifx\svgscale\undefined%
      \relax%
    \else%
      \setlength{\unitlength}{\unitlength * \real{\svgscale}}%
    \fi%
  \else%
    \setlength{\unitlength}{\svgwidth}%
  \fi%
  \global\let\svgwidth\undefined%
  \global\let\svgscale\undefined%
  \makeatother%
  \begin{picture}(1,0.48603642)%
    \put(0,0){\includegraphics[width=\unitlength,page=1]{setup.pdf}}%
    \put(0.21874525,0.46493269){\color[rgb]{0.46666667,0.57647059,0.23529412}\makebox(0,0)[lt]{\begin{minipage}{0.17910084\unitlength}\raggedright Elliptical Locii\end{minipage}}}%
    \put(0.35827271,0.16940973){\color[rgb]{0.19215686,0.34117647,0.57254902}\rotatebox{0.16123007}{\makebox(0,0)[lb]{\smash{Diffuse Wall}}}}%
    \put(0.66299217,0.36935417){\color[rgb]{0,0,0}\makebox(0,0)[lb]{\smash{$\px$}}}%
    \put(0.86501428,0.4175882){\color[rgb]{0,0,0}\makebox(0,0)[lb]{\smash{$\light$}}}%
    \put(0.86923993,0.35842932){\color[rgb]{0,0,0}\makebox(0,0)[lb]{\smash{$\vispoint_1$}}}%
    \put(0.66702687,0.4763245){\color[rgb]{0,0,0}\makebox(0,0)[lb]{\smash{$\Voxels$}}}%
    \put(0.66464673,0.11664067){\color[rgb]{0,0,0}\makebox(0,0)[lb]{\smash{$\px$}}}%
    \put(0.86501428,0.16487469){\color[rgb]{0,0,0}\makebox(0,0)[lb]{\smash{$\light$}}}%
    \put(0.86923993,0.10571582){\color[rgb]{0,0,0}\makebox(0,0)[lb]{\smash{$\vispoint_1$}}}%
    \put(0.66923298,0.22195641){\color[rgb]{0,0,0}\makebox(0,0)[lb]{\smash{$\Voxels$}}}%
    \put(0.86923993,0.04504761){\color[rgb]{0,0,0}\makebox(0,0)[lb]{\smash{$\vispoint_2$}}}%
    \put(0.12372144,0.30845323){\color[rgb]{0,0,0}\makebox(0,0)[lb]{\smash{$\px$}}}%
    \put(0.35977598,0.35809089){\color[rgb]{0,0,0}\makebox(0,0)[lb]{\smash{$\light$}}}%
    \put(0.3652913,0.28087683){\color[rgb]{0,0,0}\makebox(0,0)[lb]{\smash{$\vispoint$}}}%
    \put(0.04430126,0.3680184){\color[rgb]{0,0,0}\makebox(0,0)[lb]{\smash{$\Scene$}}}%
    \put(0.25893222,0.13668277){\color[rgb]{0,0,0}\makebox(0,0)[lb]{\smash{$\pixel$}}}%
    \put(0.04878332,0.17947853){\color[rgb]{0,0,0}\makebox(0,0)[lb]{\smash{Occluder}}}%
    \put(0.04017546,0.11416669){\color[rgb]{0,0,0}\makebox(0,0)[lb]{\smash{Laser}}}%
    \put(0.17445165,0.03602896){\color[rgb]{0,0,0}\makebox(0,0)[lb]{\smash{Observer}}}%
    \put(0.17334859,0.01176173){\color[rgb]{0,0,0}\makebox(0,0)[lb]{\smash{\relsize{-1} (camera)}}}%
  \end{picture}%
\endgroup%

%% file: gandalfUp_combined.pdf_tex
\begingroup%
  \makeatletter%
  \providecommand\color[2][]{%
    \errmessage{(Inkscape) Color is used for the text in Inkscape, but the package 'color.sty' is not loaded}%
    \renewcommand\color[2][]{}%
  }%
  \providecommand\transparent[1]{%
    \errmessage{(Inkscape) Transparency is used (non-zero) for the text in Inkscape, but the package 'transparent.sty' is not loaded}%
    \renewcommand\transparent[1]{}%
  }%
  \providecommand\rotatebox[2]{#2}%
  \ifx\svgwidth\undefined%
    \setlength{\unitlength}{2024.13702393bp}%
    \ifx\svgscale\undefined%
      \relax%
    \else%
      \setlength{\unitlength}{\unitlength * \real{\svgscale}}%
    \fi%
  \else%
    \setlength{\unitlength}{\svgwidth}%
  \fi%
  \global\let\svgwidth\undefined%
  \global\let\svgscale\undefined%
  \makeatother%
  \begin{picture}(1,0.33809165)%
    \put(0,0){\includegraphics[width=\unitlength,page=1]{gandalfUp_combined.pdf}}%
    \put(0.10091442,0.01460986){\color[rgb]{0,0,0}\makebox(0,0)[lb]{\smash{Traditional - 1873.34s }}}%
    \put(0.46185359,0.01460986){\color[rgb]{0,0,0}\makebox(0,0)[lb]{\smash{Our method - 19.37s (96.7x) }}}%
  \end{picture}%
\endgroup%

%% file: spad_clamped_combined_background_front.pdf_tex
\begingroup%
  \makeatletter%
  \providecommand\color[2][]{%
    \errmessage{(Inkscape) Color is used for the text in Inkscape, but the package 'color.sty' is not loaded}%
    \renewcommand\color[2][]{}%
  }%
  \providecommand\transparent[1]{%
    \errmessage{(Inkscape) Transparency is used (non-zero) for the text in Inkscape, but the package 'transparent.sty' is not loaded}%
    \renewcommand\transparent[1]{}%
  }%
  \providecommand\rotatebox[2]{#2}%
  \ifx\svgwidth\undefined%
    \setlength{\unitlength}{1641.74285889bp}%
    \ifx\svgscale\undefined%
      \relax%
    \else%
      \setlength{\unitlength}{\unitlength * \real{\svgscale}}%
    \fi%
  \else%
    \setlength{\unitlength}{\svgwidth}%
  \fi%
  \global\let\svgwidth\undefined%
  \global\let\svgscale\undefined%
  \makeatother%
  \begin{picture}(1,0.30018417)%
    
    \put(0,0){\includegraphics[width=\unitlength,page=1]{spad_clamped_combined_background_front.pdf}}%
    \put(0.11,.275){\color[rgb]{0,0,0}\makebox(0,0)[lb]{\smash{Traditional - 14.8s }}}%
    \put(0.44911723,.275){\color[rgb]{0,0,0}\makebox(0,0)[lb]{\smash{Our method - 1.6s (8.2x) }}}%
  \end{picture}%
\endgroup%

%% file: LOWMID_168.pdf_tex
\begingroup%
  \makeatletter%
  \providecommand\color[2][]{%
    \errmessage{(Inkscape) Color is used for the text in Inkscape, but the package 'color.sty' is not loaded}%
    \renewcommand\color[2][]{}%
  }%
  \providecommand\transparent[1]{%
    \errmessage{(Inkscape) Transparency is used (non-zero) for the text in Inkscape, but the package 'transparent.sty' is not loaded}%
    \renewcommand\transparent[1]{}%
  }%
  \providecommand\rotatebox[2]{#2}%
  \ifx\svgwidth\undefined%
    \setlength{\unitlength}{192bp}%
    \ifx\svgscale\undefined%
      \relax%
    \else%
      \setlength{\unitlength}{\unitlength * \real{\svgscale}}%
    \fi%
  \else%
    \setlength{\unitlength}{\svgwidth}%
  \fi%
  \global\let\svgwidth\undefined%
  \global\let\svgscale\undefined%
  \makeatother%
  \begin{picture}(1,1)%
    \put(0,0){\includegraphics[width=\unitlength,page=1]{LOWMID_168.pdf}}%
    \put(0.5,0.04793948){\color[rgb]{0,0,0}\makebox(0,0)[cb]{\smash{8.59s}}}%
    \put(0.5,0.86573608){\color[rgb]{0,0,0}\makebox(0,0)[cb]{\smash{Ours - 168 triang.}}}%
  \end{picture}%
\endgroup%

%% file: MID_656.pdf_tex
\begingroup%
  \makeatletter%
  \providecommand\color[2][]{%
    \errmessage{(Inkscape) Color is used for the text in Inkscape, but the package 'color.sty' is not loaded}%
    \renewcommand\color[2][]{}%
  }%
  \providecommand\transparent[1]{%
    \errmessage{(Inkscape) Transparency is used (non-zero) for the text in Inkscape, but the package 'transparent.sty' is not loaded}%
    \renewcommand\transparent[1]{}%
  }%
  \providecommand\rotatebox[2]{#2}%
  \ifx\svgwidth\undefined%
    \setlength{\unitlength}{192bp}%
    \ifx\svgscale\undefined%
      \relax%
    \else%
      \setlength{\unitlength}{\unitlength * \real{\svgscale}}%
    \fi%
  \else%
    \setlength{\unitlength}{\svgwidth}%
  \fi%
  \global\let\svgwidth\undefined%
  \global\let\svgscale\undefined%
  \makeatother%
  \begin{picture}(1,1)%
    \put(0,0){\includegraphics[width=\unitlength,page=1]{MID_656.pdf}}%
    \put(0.5,0.04793948){\color[rgb]{0,0,0}\makebox(0,0)[cb]{\smash{8.92s}}}%
    \put(0.5,0.86573608){\color[rgb]{0,0,0}\makebox(0,0)[cb]{\smash{Ours - 656 triang.}}}%
  \end{picture}%
\endgroup%

%% file: HIGH_2592.pdf_tex
\begingroup%
  \makeatletter%
  \providecommand\color[2][]{%
    \errmessage{(Inkscape) Color is used for the text in Inkscape, but the package 'color.sty' is not loaded}%
    \renewcommand\color[2][]{}%
  }%
  \providecommand\transparent[1]{%
    \errmessage{(Inkscape) Transparency is used (non-zero) for the text in Inkscape, but the package 'transparent.sty' is not loaded}%
    \renewcommand\transparent[1]{}%
  }%
  \providecommand\rotatebox[2]{#2}%
  \ifx\svgwidth\undefined%
    \setlength{\unitlength}{192bp}%
    \ifx\svgscale\undefined%
      \relax%
    \else%
      \setlength{\unitlength}{\unitlength * \real{\svgscale}}%
    \fi%
  \else%
    \setlength{\unitlength}{\svgwidth}%
  \fi%
  \global\let\svgwidth\undefined%
  \global\let\svgscale\undefined%
  \makeatother%
  \begin{picture}(1,1)%
    \put(0,0){\includegraphics[width=\unitlength,page=1]{HIGH_2592.pdf}}%
        \put(0.5,0.04793948){\color[rgb]{0,0,0}\makebox(0,0)[cb]{\smash{9.91s}}}%
    \put(0.5,0.86573608){\color[rgb]{0,0,0}\makebox(0,0)[cb]{\smash{Ours - 2592 triang.}}}%
  \end{picture}%
\endgroup%

%% file: CPU.pdf_tex
\begingroup%
  \makeatletter%
  \providecommand\color[2][]{%
    \errmessage{(Inkscape) Color is used for the text in Inkscape, but the package 'color.sty' is not loaded}%
    \renewcommand\color[2][]{}%
  }%
  \providecommand\transparent[1]{%
    \errmessage{(Inkscape) Transparency is used (non-zero) for the text in Inkscape, but the package 'transparent.sty' is not loaded}%
    \renewcommand\transparent[1]{}%
  }%
  \providecommand\rotatebox[2]{#2}%
  \ifx\svgwidth\undefined%
    \setlength{\unitlength}{192bp}%
    \ifx\svgscale\undefined%
      \relax%
    \else%
      \setlength{\unitlength}{\unitlength * \real{\svgscale}}%
    \fi%
  \else%
    \setlength{\unitlength}{\svgwidth}%
  \fi%
  \global\let\svgwidth\undefined%
  \global\let\svgscale\undefined%
  \makeatother%
  \begin{picture}(1,1)%
    \put(0,0){\includegraphics[width=\unitlength,page=1]{CPU.pdf}}%
    \put(0.5,0.04793948){\color[rgb]{0,0,0}\makebox(0,0)[cb]{\smash{13167.44s}}}%
    \put(0.5,0.86573608){\color[rgb]{0,0,0}\makebox(0,0)[cb]{\smash{Traditional}}}%
  \end{picture}%
\endgroup%